\documentclass[11pt,a4paper]{article}
\usepackage[hyperref]{style/emnlp2020}
\usepackage{times}
\usepackage{latexsym}

\usepackage{microtype}

\aclfinalcopy %
\def\aclpaperid{361} %

\title{Understanding tables with intermediate pre-training}

\author{Julian Martin Eisenschlos, Syrine Krichene, Thomas M{\"u}ller \\ \\
  Google Research, Z{\"u}rich \\
  \texttt{\{eisenjulian,syrinekrichene,thomasmueller\}@google.com}}

\date{}

\usepackage{amsthm}

\usepackage{framed}
\usepackage{mdwlist}
\usepackage{colortbl}
\usepackage{xcolor}
\usepackage{nicefrac}
\usepackage{booktabs}
\usepackage{amsfonts}
\usepackage[T1]{fontenc}
\usepackage{bold-extra}
\usepackage{amsmath}
\usepackage{amssymb}
\usepackage{bm}
\usepackage{graphicx}
\usepackage{mathtools}
\usepackage{multirow}
\usepackage{multicol}
\usepackage[normalem]{ulem}
\usepackage{lipsum}
\usepackage{float}
\usepackage{ifthen}
\usepackage[ruled,vlined]{algorithm2e}

\usepackage{latexsym,comment}
\usepackage{tikz}
\usepackage{xspace}
\usepackage{xr}
\usetikzlibrary{shapes,arrows}

\newcommand{\gem}[1]{\mbox{\textsc{gem}}}
\newcommand{\abr}[1]{\textsc{#1}}

\DeclareMathOperator*{\argmax}{arg\,max}

\newcommand{\hidetext}[1]{}
\newcommand{\ignore}[1]{}

\newif\ifcomment\commenttrue

\newif\ifinfotabs\infotabsfalse

\newif\ifsubscripterror\subscripterrortrue

\ifcomment
\newcommand{\pinaforecomment}[3]{\colorbox{#1}{\parbox{.8\linewidth}{#2: #3}}}
\else
\newcommand{\pinaforecomment}[3]{}
\fi

\ifinfotabs
\newcommand{\infotabstext}[1]{#1}
\else
\newcommand{\infotabstext}[1]{}
\fi

\ifsubscripterror
\newcommand{\err}[1]{\textsubscript{~$\pm$#1}}
\else
\newcommand{\err}[1]{ $\pm$ #1}
\fi

\newcommand{\smallurl}[1]{ \begin{tiny}\url{#1}\end{tiny}}

\definecolor{lightblue}{HTML}{3cc7ea}
\definecolor{CUgold}{HTML}{CFB87C}
\definecolor{grey}{rgb}{0.95,0.95,0.95}
\definecolor{ceil}{rgb}{0.57, 0.63, 0.81}
\definecolor{UMDred}{HTML}{ed1c24}
\definecolor{UMDyellow}{HTML}{ffc20e}
\definecolor{darkgreen}{HTML}{008f00}

\newcommand{\sqa}{\textsc{SQA}\xspace}
\newcommand{\tabfact}{\textsc{TabFact}\xspace}
\newcommand{\boolq}{\textsc{BoolQ}\xspace}
\newcommand{\infotabs}{\textsc{InfoTabs}\xspace}

\newcommand{\editout}[1]{{\color{red} \sout{#1}}}
\newcommand{\editin}[1]{{\color{darkgreen} \textbf{#1}}}

\newcommand{\cls}{\texttt{[CLS]}}
\newcommand{\sep}{\texttt{[SEP]}}
\newcommand\ours{\textsc{OURS}\xspace}

\newcommand\sota{state-of-the-art\xspace}
\newcommand{\tapas}{\textsc{TAPAS}\xspace}
\newcommand{\bert}{\textsc{BERT}\xspace}
\newcommand{\roberta}{\textsc{RoBERTa}\xspace}
\newcommand{\tablebert}{\textsc{Table-BERT}\xspace}
\newcommand{\lfc}{\textsc{LogicalFactChecker}\xspace}
\newcommand{\masklm}{\textsc{Mask-LM}\xspace}

\newcommand\la{$\langle$}  %
\newcommand\ra{$\rangle$}
\newcommand{\bftab}{\fontseries{b}\selectfont}

\theoremstyle{definition}

\newcommand{\errordescr}{Error margins are estimated as half the interquartile range}

\newcommand{\pruningtable}[1]{
\resizebox{0.8\columnwidth}{!}{
\begin{tabular}{lcll}
\toprule
\textbf{Method} & \textbf{PT Size} & \textbf{FT Size} & \multicolumn{1}{l}{\textbf{Val}} \\ 
\midrule
\tablebert{} &  & 512\ifthenelse{\equal{#1}{full}}{}{\footnotemark} & 66.1 \\ 
\midrule
\ours & 512 & 512 & 78.3\err{0.2} \\
      & 256 & 512 &	78.6\err{0.3} \\
      & 128 & 512 &	77.5\err{0.3} \\
\midrule
\ours{} - HEL  & 128  & 512 & 76.7\err{0.4} \\
               & 128  & 256 & 76.3\err{0.1} \\
               & 128  & 128 & 71.0\err{0.3} \\
\midrule
\ours{} - HEM &  256 & 512 & 78.8\err{0.3} \\
              &  256 & 256 & 78.1\err{0.1} \\
              &  128 & 512 & 78.2\err{0.4} \\
              &  128 & 256 & 77.0\err{0.2} \\
\ifthenelse{\equal{#1}{full}}{
    &  128 & 128 & 72.7\err{0.2} \\
\midrule
\ours - W2V	& 128 & 512 & 77.7\err{0.3} \\
	        & 128 & 256 & 76.0\err{0.2} \\
	        & 128 & 128 & 70.6\err{0.3} \\
\midrule
\ours - IWF & 128 & 512 & 77.9\err{0.2} \\
            & 128 & 256 & 77.2\err{0.1} \\
            & 128 & 128 & 72.7\err{0.3} \\
\midrule            
\ours - CHAR & 128 & 512 & 77.5\err{0.2} \\
             & 128 & 256 & 74.8\err{0.1} \\
             & 128 & 128 & 68.7\err{0.0} \\
}{
    &  128 & 128 & 72.7\err{0.2} \\
}
\bottomrule
\end{tabular}}%
\ifthenelse{\equal{#1}{full}}{
\caption{Accuracy of different pruning methods: The heuristic entity linking (HEL) \cite{2019TabFactA}, Heuristic exact match (HEM), word-to-vec (W2V), inverse word frequency (IWF), character ngram (CHAR) at different pre-training (PT) and fine-tuning (FT) sizes. \errordescr.}
}{
\caption{Accuracy of column pruning methods, that reduce input length for faster training and prediction: The heuristic entity linking (HEL) \cite{2019TabFactA} and Heuristic exact match (HEM) at various pre-training (PT) and fine-tuning (FT) sizes. HEM out-performs HEL on all input sizes, and in the faster case (128) out-performs \tablebert by 6.6 points. Accuracy with size 256 is 0.7 points behind the full input size. \errordescr.}
}
} 

\makeatletter
\newcommand*{\addFileDependency}[1]{%
  \typeout{(#1)}
  \@addtofilelist{#1}
  \IfFileExists{#1}{}{\typeout{No file #1.}}
}
\makeatother

\begin{document}
\maketitle

\begin{abstract}

Table entailment, the binary classification task of finding if a sentence is supported or refuted by the content of a table, requires parsing language and table structure as well as numerical and discrete reasoning.
While there is extensive work on textual entailment, table entailment is less well studied.
We adapt \tapas~\cite{herzig-2020}, a table-based \bert model, to recognize entailment. 
Motivated by the benefits of data augmentation, we create a balanced dataset of millions of automatically created training examples which are learned in an intermediate step prior to fine-tuning.
This new data is not only useful for table entailment, but also for \sqa~\cite{iyyer-etal-2017-search}, a sequential table QA task.
To be able to use long examples as input of \bert~models, we evaluate table pruning techniques as a pre-processing step to drastically improve the training and prediction efficiency at a moderate drop in accuracy.
The different methods set the new \sota on the \tabfact~\cite{2019TabFactA} and \sqa~datasets.
\end{abstract}

\section{Introduction}
\label{sec:intro}

\begin{figure*}[!t]
\centering
\begin{tabular}{llllll}
\textbf{Rank} &	\textbf{Player} & \textbf{Country} & \textbf{Earnings} & \textbf{Events} & \textbf{Wins} \\
\hline
1	&Greg Norman    &	Australia    &	1,654,959&	16&	3 \\
2	&Billy Mayfair  &	United States&	1,543,192&	28&	2 \\
3	&Lee Janzen     &	United States&	1,378,966&	28&	3 \\
4	&Corey Pavin    &	United States&	1,340,079&	22&	2 \\
5	&Steve Elkington&	Australia    &	1,254,352&	21&	2 \\
\end{tabular}%
\\
\vspace{12pt}
\begin{tabular}{ll}
{\small \emph{Entailed:}} &{\small Greg Norman and Steve Elkington are from the same country. }\\
                         &{\small Greg Norman and Lee Janzen both have 3 wins.}\\

{\small \emph{Refuted:}}  &{\small Greg Norman is from the US and Steve Elkington is from Australia. }\\
                         &{\small Greg Norman and Billy Mayfair tie in rank.}\\

{\small \emph{Counterfactual:}} & {\small \editin{Greg Norman} has the highest earnings.}\\
                                & {\small \editout{Steve Elkington} has the highest earnings.}\\
{\small\emph{Synthetic:}}&{\small $2$ is less than wins when Player is Lee Janzen.}\\
&{\small The sum of Earnings when Country is Australia is $2,909,311$.}\\
\end{tabular}%
\caption{A \tabfact{} table with real statements\footnotemark and counterfactual and synthetic examples.}
\label{fig:example}
\end{figure*}

Textual entailment~\cite{dagan2005pascal}, also known as natural language inference~\cite{bowman-etal-2015-large}, is a core natural language processing (\abr{nlp}) task.
It can predict effectiveness of reading comprehension~\cite{dagan-10}, which argues that it can form the foundation of many other \abr{nlp} tasks, and is a useful neural pre-training task~\cite{subramanian2018learning,conneau-etal-2017-supervised}.

Textual entailment is well studied, but many relevant data sources are structured or semi-structured: 
health data both worldwide and personal, fitness trackers, stock markets, and sport statistics.
While some information needs can be anticipated by hand-crafted templates, user queries are often surprising, and having models that can reason and parse that structure can have a great impact in real world applications~\cite{Khashabi19, aristo}.

A recent example is \tabfact~\cite{2019TabFactA}, a dataset of statements that are either entailed or refuted by tables from Wikipedia (Figure~\ref{fig:example}).
Because solving these entailment problems requires sophisticated reasoning and higher-order operations like $\argmax$, averaging, or comparing, human accuracy remains substantially (18 points) ahead of the best models~\cite{zhong2020logicalfactchecker}.

The current models are dominated by semantic parsing approaches that attempt to create logical forms from weak supervision.
We, on the other hand, follow \citet{herzig-2020} and \citet{2019TabFactA} and encode the tables with \bert{}-based models to directly predict the entailment decision.
But while \bert{} models for text have been scrutinized and optimized for how to best pre-train and represent \emph{textual} data, the same attention has not been applied to tabular data, limiting the effectiveness in this setting.
This paper addresses these shortcomings using \emph{intermediate task} pre-training~\cite{pruksachatkun-2020-intermediate-task}, creating efficient data representations, and applying these improvements to the tabular entailment task. 

Our methods are tested on the English language, mainly due to the availability of the end task resources. However, we believe that the proposed solutions could be applied in other languages where a pre-training corpus of text and tables is available, such as the Wikipedia datasets.

Our main contributions are the following: 

i) We introduce two \emph{intermediate} pre-training tasks, which are learned from a trained \masklm{}~model, one based on synthetic and the other from counterfactual statements.
The first one generates a sentence by sampling from a set of logical expressions that filter, combine and compare the information on the table, which is required in table entailment (e.g., knowing that Gerald Ford is taller than the average president requires summing all presidents and dividing by the number of presidents).
The second one corrupts sentences about tables appearing on Wikipedia by swapping entities for plausible alternatives.
Examples of the two tasks can be seen in Figure \ref{fig:example}. 
The procedure is described in detail in section \ref{sec:methods}.

ii)
We demonstrate column pruning to be an effective means of lowering computational cost at minor drops in accuracy, doubling the inference speed at the cost of less than one accuracy point.

iii) Using the pre-training tasks, we set the new \sota on \tabfact{} out-performing previous models by $6$ points when using a \bert{}-base model and $9$ points for a \bert{}-large model. The procedure is data efficient and can get comparable accuracies to previous approaches when using only 10\% of the data.
We perform a detailed analysis of the improvements in Section \ref{sec:analysis}.
Finally, we show that our method improves the \sota on a question answering task (\sqa{}) by 4 points.

We release the pre-training checkpoints, data generation and training code at \href{https://github.com/google-research/tapas}{github.com/google-research/tapas}.
\footnotetext{Based on table \texttt{2-14611590-3.html} with light edits.}

\section{Model}
\label{sec:model}

We use a model architecture derived from \bert{} and add additional embeddings to encode the table structure, following the approach of \citet{herzig-2020} to encode the input. 

The statement and table in a pair are tokenized into word pieces and concatenated using the standard \texttt{[CLS]} and \texttt{[SEP]} tokens in between. The table is flattened row by row and no additional separator is added between the cells or rows.

Six types of learnable input embeddings are added together as shown in Appendix \ref{sec:apx-input}. 
\textbf{Token embeddings}, \textbf{position embeddings} and \textbf{segment embeddings} are analogous to the ones used in standard \bert{}. 
Additionally we follow \citet{herzig-2020} and use \textbf{column and row embeddings} which encode the two dimensional position of the cell that the token corresponds to and
\textbf{rank embeddings} for numeric columns that encode the numeric rank of the cell with respect to the column, and provide a simple way for the model to know how a row is ranked according to a specific column.

Recall that the bi-directional self-attention mechanism in transformers is unaware of order, which motivates the usage of positional and segment embeddings for text in \bert{}, and generalizes naturally to column and row embeddings when processing tables, in the $2$-dimensional case.

Let $s$ and $T$ represent the sentence and table respectively and $E_s$ and $E_T$ be their corresponding input embeddings. 
The sequence $E = [E_\cls; E_s; E_\sep; E_T]$ is passed through a transformer~\cite{vaswani-2017} denoted $f$ and a contextual representation is obtained for every token. We model the probability of entailment $P(s|T)$ with a single hidden layer neural network computed from the output of the \cls ~token:
\[
P(s|T) = \texttt{MLP}\big(f_\cls\left(E\right)\big)
\]
where the middle layer has the same size as the hidden dimension and uses a \texttt{tanh} activation and the final layer uses a \texttt{sigmoid} activation.

\section{Methods}
\label{sec:methods}

The use of challenging pre-training tasks has been successful in improving downstream accuracy~\cite{Clark2020ELECTRA:}.
One clear caveat of the method adopted in \citet{herzig-2020} which attempts to fill in the blanks of sentences and cells in the table is that not much understanding of the table in relation with the sentence is needed. 

With that in mind, we propose two tasks that require sentence-table reasoning and feature complex operations performed on the table and entities grounded in sentences in non-trivial forms.

We discuss two methods to create pre-training data that lead to stronger table entailment models.
Both methods create statements for existing Wikipedia tables\footnote{Extracted from a Wikipedia dump from 12-2019.}. 
We extract all tables that have at least two columns, a header row and two data rows.
We recursively split tables row-wise into the upper and lower half until they have at most 50 cells.
This way we obtain 3.7 million tables.

\subsection{Counterfactual Statements}

Motivated by work on counterfactually-augmented data ~\cite{Kaushik2020Learning, gardner-2020},
we propose an automated and scalable method to get table entailments from Wikipedia and, for each such positive examples, create a minimally differing refuted example.
For this pair to be useful we want that their truth value can be predicted from the associated table but not without it.

The tables and sentences are extracted from Wikipedia as follows:
We use the page title, description, section title, text and caption. 
We also use all sentences on Wikipedia that link to the table's page and mentions at least one page (entity) that is also mentioned in the table.
Then these snippets are split into sentences using the NLTK~\cite{loper-02} implementation of Punkt~\cite{kiss-06}.
For each relevant sentence we create one positive and one negative statement.

Consider the table in Figure \ref{fig:example} and the sentence \emph{`[Greg Norman] is [Australian].'} (Square brackets indicate mention boundaries.).
A mention\footnote{We annotate numbers and dates in the table and sentence with a simple parser and rely on the Wikipedia mention annotations (anchors) for identifying entities.} is a potential \textbf{focus mention} if the same entity or value is also mentioned in the table. In our example, \emph{Greg Norman} and \emph{Australian} are potential focus mentions.
Given a focus mention (\emph{Greg Norman}) we define all the mentions that occur in the same column (but do not refer to the same entity) as the \textbf{replacement mentions} (e.g., \emph{Billy Mayfair}, \emph{Lee Janzen}, \dots). We expect to create a false statement if we replace the focus mention with a replacement mention (e.g., \emph{`Billy Mayfair is Australian.'}), but there is no guarantee it will be actually false.

We call a mention of an entity that occurs in the same row as the focus entity a \textbf{supporting mention}, because it increases the chance that we falsify the statement by replacing the focus entity. In our example, \emph{Australian} would be a supporting mention for \emph{Greg Norman} (and vice versa). If we find a supporting mention we restrict the replacement candidates to the ones that have a different value. In the example, we would not use \emph{Steve Elkington} since his row also refers to Australia.

Some replacements can lead to ungrammatical statements that a model could use to identify the negative statements,
so we found it is useful to also replace the entity in the original positive sentence from Wikipedia with the mention from the table.\footnote{
 Consider that if \emph{Australian} is our focus and we replace it with \emph{United States} we get `\emph{Greg Norman is United States.}'.} \\
We also introduce a simple type system for entities (named entity, date, cardinal number and ordinal number) and only replace entities of the same type.
Short sentences having less than $4$ tokens not counting the mention, are filtered out.

Using this approach we extract 4.1 million counterfactual pairs of which 546 thousand do have a supporting mention and the remaining do not.

We evaluated 100 random examples manually and found that the percentage of negative statements that are false and can be refuted by the table is 82\% when they have a supporting mention and 22\% otherwise.
Despite this low value we still found the examples without supporting mention to improve accuracy on the end tasks (Appendix \ref{sec:apx-table-pruning}).

\subsection{Synthetic Statements}

\begin{figure}[!t]
\centering
\begin{tabular}{rl}
\la statement\ra $~~\to$ & \la expr\ra \la compare\ra \la expr\ra \\
\la expr\ra $~~\to$ & \la select\ra ~when \la where\ra ~| \\ & \la select\ra \\
\la select\ra $~~\to$ &  \la column\ra ~| \\ & the \la aggr\ra ~of \la column\ra ~| \\ & the count \\
\la where\ra $~~\to$ & \la column\ra \la compare\ra \la value\ra ~|\\ & \la where\ra ~and \la where\ra \\
\la aggr\ra$~~\to$ & first | last | \\ & lowest | greatest | \\ & sum | average | range \\
\la compare\ra $~~\to$ & is | \\ & is greater than | \\ & is less than \\
\la value\ra  $~~\to$ & \la string\ra ~| \la number\ra
\end{tabular}
\caption{Grammar of synthetic phrases. \la column\ra ~is the set of column names in the table. We also generate constant expressions by replacing expressions with their values. Aggregations are defined in Table \ref{table:aggregations}.}
\label{fig:grammar}
\end{figure}

\begin{table}[!t]
\centering
\resizebox{1\columnwidth}{!}{
\begin{tabular}{ll}
\textbf{Name}        & \textbf{Result} \\ 
\hline
first                & the value in C with the lowest row index. \\
last                 & the value in C with the highest row index. \\
greatest             & the value in C with the highest numeric value. \\
lowest               & the value in C with the lowest numeric value. \\
sum                  & The sum of all the numeric values.\\
average              & The average of all the numeric values.\\
range                & The difference between greatest and lowest.\\
\hline
\end{tabular}
\vspace{-12pt}
}
\caption{Aggregations used in synthetic statements, where $C$ are the column values. When $C$ is empty or a singleton, it results in an error. Numeric functions also fail if any of their values is non-numeric.}
\label{table:aggregations}
\end{table}

Motivated by previous work~\cite{Geva2020InjectingNR}, we propose a synthetic data generation method to improve the handling of numerical operations and comparisons.
We build a table-dependent \emph{statement} that compares two simplified SQL-like expressions.
We define the (probabilistic) \emph{context-free grammar} shown in Figure \ref{fig:grammar}.
Synthetic statements are sampled from the CFG.
We constrain the  \la select\ra~values of the left and right expression to be either both \emph{the count} or to have the same value for \la column\ra.
This guarantees that the domains of both expressions are comparable.
\la value\ra~is chosen as at random from the respective column.
A statement is redrawn if it yields an error (see Table \ref{table:aggregations}).

With probability 0.5 we replace one of both expressions by the values it evaluates to.
In the example given in figure \ref{fig:example}, ``[The [sum] of [Earnings]] when [[Country] [is] [Australia]]'' is an \la expr\ra~that can be replaced by the constant value $2,909,311$.

We set $P(\mbox{\la select\ra} \to \mbox{the count})$ to $0.2$ in all our experiments. Everything else is sampled uniformly. 
For each Wikipedia table we generate a positive and a negative statement which yields $3.7$M pairs.

\subsection{Table pruning}
\label{sec:pruning}
Some input examples from \tabfact{} can be too long for \bert-based models. We evaluate table pruning techniques as a pre-processing step to select relevant columns that respect the input example length limits.
As described in section~\ref{sec:model}, an example is built by concatenating the statement with the flattened table. 
For large tables the example length can exceed the capacity limit of the transformer.

The \tapas{} model handles this by shrinking the text in cells. A token selection algorithm loops over the cells. For each cell it starts by selecting the first token, then the second and so on until the maximal length is reached. 
Unless stated otherwise we use the same approach.
Crucially, selecting only relevant columns would allow longer examples to fit without discarding potentially relevant tokens.

\textbf{Heuristic entity linking (HEL)} is used as a baseline. It is the table pruning used in \tablebert{}~\cite{2019TabFactA}. 
The algorithm aligns spans in statement to the columns
by extracting the longest character n-gram that matches a cell. %
The span matches represent linked entities. 
Each entity in the statement can be linked to only one column.
We use the provided entity linking statements data\footnote{\href{https://github.com/wenhuchen/Table-Fact-Checking/blob/master/tokenized_data}{github.com/wenhuchen/Table-Fact-Checking/\\blob/master/tokenized\_data}}.
We run the \tapas{} algorithm on top of the input data to limit the input size.

We propose a different method that tries to retain as many columns as possible.
In our method, the columns are ranked by a relevance score and added in order of decreasing relevance.
Columns that exceed the maximum input length are skipped.
The algorithm is detailed in Appendix \ref{sec:apx-table-pruning}.
\textbf{Heuristic exact match (HEM)} computes the Jaccard coefficient between the statement and each column.
Let $T_S$ be the set of tokens in the statement $S$ and $T_C$ the tokens in column $C$, with $C \in \mathbb{C}$ the set of columns. Then the score between the statement and column is given by $\frac{|T_S \cap T_C|}{|T_S \cup T_C|}$.

We also experimented with approaches based on word2vec \cite{mikolov-13}, character overlap and TF-IDF.
Generally, they produced worse results than HEM.
Details are shown in Appendix~\ref{sec:apx-table-pruning}.

\section{Experimental Setup}
\label{sec:experiments}

In all experiments, we start with the public \tapas{} checkpoint,\footnote{\href{https://github.com/google-research/tapas}{github.com/google-research/tapas}} train an entailment model on the data from Section \ref{sec:methods} and then fine-tune on the end task (\tabfact or \sqa). We report the median accuracy values over $3$ pre-training and $3$ fine-tuning runs ($9$ runs in total). 
We estimate the error margin as half the \emph{interquartile range}, that is half the difference between the 25th and 75th percentiles.
The hyper-parameters, how we chose them, hardware and other information to reproduce our experiments are explained in detail in Appendix~\ref{sec:apx-repro}.

The training time depends on the sequence length used. For a \bert{}-Base model it takes around $78$ minutes using $128$ tokens and it scales almost linearly up to $512$. For our pre-training tasks, we explore multiple lengths and how they trade-off speed for downstream results.

\subsection{Datasets}

We evaluate our model on the recently released \tabfact~dataset \cite{2019TabFactA}. 
The tables are extracted from Wikipedia and the sentences written by crowd workers in two batches. The first batch consisted of \textbf{simple} sentences, that instructed the writers to refer to a single row in the table. The second one, created \textbf{complex} sentences by asking writers to use information from multiple rows.

In both cases, crowd workers initially created only positive (entailed) pairs, and in a subsequent annotation job, the sentences were copied and edited into negative ones, with instructions of avoiding simple negations. Finally, there was a third verification step to filter out bad rewrites.
The final count is $118,000$. The split sizes are given in Appendix ~\ref{sec:apx-dataset-stats}.
An example of a table and the sentences is shown in Figure~\ref{fig:example}. 
We use the standard \tabfact split and the official accuracy metric.

\infotabstext{
\infotabs{}~\cite{gupta2020infotabs} is another recently released table entailment dataset. It differs from \tabfact in that it only uses Wikipedia info-boxes. 
}

We also use the \sqa{} \cite{iyyer-etal-2017-search} dataset for pre-training (following \citet{herzig-2020}) and for testing if our pre-training is useful for related tasks.
\sqa is a question answering dataset that was created by asking crowd workers to split a compositional subset of WikiTableQuestions \cite{pasupat2015compositional} into multiple referential questions.
The dataset consists of 6,066 sequences ($2.9$ question per sequence on average).
We use the standard split and official evaluation script.

\subsection{Baselines}

\citet{2019TabFactA} present two models, \tablebert and the Latent Program Algorithm (LPA), that yield similar accuracy on the \tabfact data.

LPA tries to predict a latent program that is then executed to verify if the statement is correct or false. The search over programs is restricted using lexical heuristics. Each program and sentence is encoded with an independent transformer model and then a linear layer gives a relevance score to the pair. The model is trained with weak supervision where programs that give the correct binary answer are considered positive and the rest negative.

\tablebert is a \bert{}-base model that similar to our approach directly predicts the truth value of the statement.
However, the model does not use special embeddings to encode the table structure but relies on a template approach to format the table as natural language. The table is mapped into a single sequence of the form: \emph{``Row 1 Rank is 1; the Player is Greg Norman; ... . Row 2 ...''}.
The model is also not pre-trained on table data.

\lfc~\cite{zhong2020logicalfactchecker} is another transformer-based model that given a candidate logical expression, combines contextual embeddings of program, sentence and table, with a tree-RNN~\cite{socher-etal-2013-parsing} to encode the parse tree of the expression. The programs are obtained through either LPA or an LSTM generator (Seq2Action).

\section{Results}

\begin{table*}[tb]
\small
\centering
\resizebox{2\columnwidth}{!}{
\begin{tabular}{llllllll} 
\toprule
\multicolumn{3}{l}{\textbf{Model}} & \multicolumn{1}{l}{\textbf{Val}} & \multicolumn{1}{l}{\textbf{Test}} & \textbf{Test\textsubscript{simple}} & \textbf{Test\textsubscript{complex}} & \textbf{Test\textsubscript{small}}  \\ 
\midrule
\multicolumn{3}{l}{\bert~classifier w/o Table}                  & 50.9                    & 50.5 & 51.0 &      50.1               & 50.4        \\ 
\midrule
\multicolumn{3}{l}{\tablebert-Horizontal-T+F-Template}  & 66.1  & 65.1 & 79.1 &  58.2 & 68.1      \\ 
\multicolumn{3}{l}{LPA-Ranking w/ Discriminator (Caption)}  & 65.1                    & 65.3                 &  78.7 &    58.5      & 68.9         \\ 
\multicolumn{3}{l}{\lfc~(program from LPA)}         & 71.7 & 71.6 & 85.5 &64.8 &74.2 \\
\multicolumn{3}{l}{\lfc~(program from Seq2Action)}  & 71.8 & 71.7 & 85.4 &65.1 & 74.3\\
\midrule
\ours &Base & \masklm{}                            & 69.6\err{4.4} & 69.9\err{3.8} & 82.0\err{5.9} & 63.9\err{2.8} & 72.2\err{4.7} \\
\ours &Base & SQA                                  & 74.9\err{0.2} & 74.6\err{0.2} & 87.2\err{0.2} & 68.4\err{0.4} & 77.3\err{0.3} \\
\ours &Base & Counterfactual                       & 75.5\err{0.5} & 75.2\err{0.4} & 87.8\err{0.4} & 68.9\err{0.5} & 77.4\err{0.3} \\
\ours &Base & Synthetic                            & 77.6\err{0.2} & 77.9\err{0.3} & 89.7\err{0.4} & 72.0\err{0.2} & 80.4\err{0.2} \\
\ours &Base & Counterfactual + Synthetic           & {\bftab 78.6}\err{0.3} & {\bftab 78.5}\err{0.3} & {\bftab 90.5}\err{0.4} & {\bftab 72.5}\err{0.3} & {\bftab 81.0}\err{0.3} \\
\midrule
\ours &Large & Counterfactual + Synthetic          & {\bftab 81.0}\err{0.1} & {\bftab 81.0}\err{0.1} & {\bftab 92.3}\err{0.3} & {\bftab 75.6}\err{0.1} & {\bftab 83.9}\err{0.3} \\
\midrule
\multicolumn{3}{l}{Human Performance}              &  -                      &      -  &  - & -      & 92.1         \\
\bottomrule
\end{tabular}
}
\caption{
The \tabfact~results. Baseline and human results are taken from \citet{2019TabFactA} and \citet{zhong2020logicalfactchecker}. 
The best \bert{}-base model while comparable in parameters out-performs \tablebert by more than 12 points.
Pre-training with counterfactual and synthetic data gives an accuracy 8 points higher than only using \masklm{} and more than 3 points higher than using \sqa.
Both counterfactual and synthetic data out-perform pre-training with a \masklm{} objective and \sqa. Joining the two datasets gives an additional improvement. 
\errordescr.}
\label{tab:results}
\end{table*}

\paragraph{\tabfact{}}

In Table \ref{tab:results} we find that our approach outperforms the previous \sota{} on \tabfact{} by more than $6$ points (Base) or more than $9$ points (Large).
A model initialized only with the public \tapas{}~\masklm{}~ checkpoint is behind \sota{} by 2 points ($71.7\%$ vs $69.9\%$).
If we train on the counterfactual data, it out-performs \lfc{} and reaches $75.2\%$ test accuracy ($+5.3$), slightly above using \sqa{}. 
Only using the synthetic data is better ($77.9\%$),
and when using both datasets it achieves $78.5\%$.
Switching from \bert{}-Base to Large improves the accuracy by another $2.5$ points.
The improvements are consistent across all test sets.

\paragraph{Zero-Shot Accuracy and low resource regimes}

The pre-trained models are in principle already complete table entailment predictors.
Therefore it is interesting to look at their accuracy on the \tabfact{} evaluation set before fine-tuning them. 
We find that the best model trained on all the pre-training data is only two points behind the fully trained \tablebert ($63.8\%$ vs $66.1\%$).
This relatively good accuracy mostly stems from the counterfactual data.

\begin{figure}[!t]
\includegraphics[width=1\linewidth]{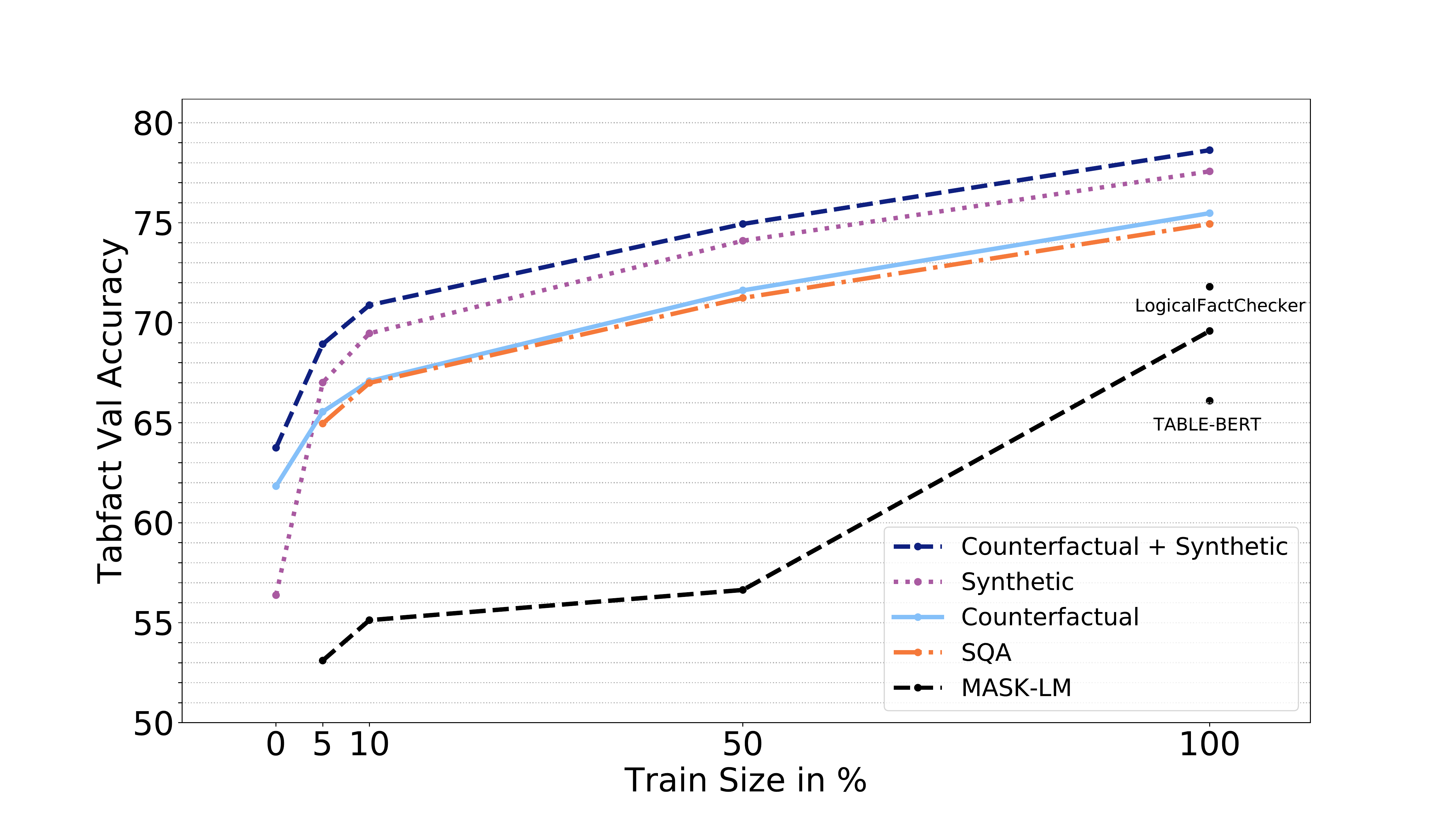}
\caption{Results for training on a subset of the data. Counterfactual + Synthetic (C+S) consistently out-performs only Counterfactual (C) or Synthetic (S), which in turn out-perform pre-training on \sqa. C+S and S surpass \tablebert{} at 5\% (around 4,500) of examples, C and \sqa at 10\%.
C+S is comparable with \lfc when using 10\% of the data. %
}
\label{fig:result-train-size}
\end{figure}

When looking at \textbf{low data regimes} in Figure \ref{fig:result-train-size} we find that
pre-training on \sqa or our artificial data consistently leads to better results than just training with the \masklm{} objective.
The models with synthetic pre-training data start out-performing \tablebert when using $5\%$ of the training set.
The setup with all the data is consistently better than the others and synthetic and counterfactual are both better than \sqa.

\infotabstext{
\paragraph{\infotabs{}} 

Results are shown in table~\ref{tab:infotabs_results}.

\begin{table}[!t]
\begin{center}
\resizebox{1.0\columnwidth}{!}{
\begin{tabular}{llrrrr}
\toprule
\textbf{Model} & \textbf{Size} & \textbf{Val} & \bm{$\alpha_1$} & \bm{$\alpha_2$} & \bm{$\alpha_3$} \\
\midrule
\citet{gupta2020infotabs}
& \bert base w/o Table & 62.69 & 63.45 & 49.65 & 50.45 \\
& \bert     base & 63.67 & 64.04 & 53.59 & 49.05\\
& \roberta  base & 68.06 & 66.70 & 56.87 & 55.26\\
& \roberta large & 77.61 & 75.06 & 69.02 & 64.61\\
\midrule
\masklm{}     & \bert base & 69.50 &  69.78 &  57.06 & 55.22 \\
Synthetic     & \bert base & 69.94 &  69.94 &  56.83 & 54.72 \\
Counterfactual& \bert base & 70.11 &  70.44 &  57.56 & 54.83 \\
Counterfactual + Synthetic
& \bert     base & 71.22 &  71.39 &  58.89 & 56.11 \\
& \bert     large& 74.44 &  74.83 & 64.11 & 61.89 \\  %
\bottomrule
\end{tabular}}
\end{center}
\caption{Results on \infotabs{}. $\alpha_1$ is a standard held-out test set, $\alpha_2$ is an adversarial set created by perturbing the statements in $\alpha_1$, $\alpha_3$ contains out-of-domain tables.}
\label{tab:infotabs_results}
\end{table}
}

\paragraph{\sqa{}} 

Our pre-training data also improves the accuracy on a QA task.
On \sqa{}~\cite{iyyer-etal-2017-search} a model pre-trained on the synthetic entailment data outperforms one pre-trained on the \masklm{}~task alone (Table \ref{tab:sqa_results}).
Our best \bert{}~Base model out-peforms the \bert{}-Large model of \citet{herzig-2020} and a \bert{}-Large model trained on our data improves the previous \sota{} by 4 points on average question and sequence accuracy. See dev results and error bars in Appendix \ref{sec:sqa_dev_results}.

\begin{table}[!t]
\begin{center}
\resizebox{1.0\columnwidth}{!}{
\begin{tabular}{lllllll}
\toprule
\textbf{Data} & \textbf{Size} & \textbf{ALL} & \textbf{SEQ}\\
\midrule
  \citet{iyyer-etal-2017-search}    &                    & 44.7  & 12.8\\
  \citet{mueller-2019}              &                    & 55.1  & 28.1\\
\citet{herzig-2020}                 &  Large             & 67.2  & 40.4 \\
\midrule
\masklm{}                        & Base  & 64.0\err{0.2} & 34.6\err{0.0} \\
Counterfactual                   & Base  & 65.0\err{0.5} & 36.5\err{0.6} \\
Synthetic                        & Base  & 67.4\err{0.2} & 39.8\err{0.4} \\
Counterf. + Synthetic            & Base  & 67.9\err{0.3} & 40.5\err{0.7} \\
\midrule
Counterf. + Synthetic            & Large & {\bftab 71.0}\err{0.4} & {\bftab 44.8}\err{0.8} &\\
\bottomrule
\end{tabular}}
\end{center}
\caption{
\sqa{} test results. ALL is the average question accuracy and SEQ the sequence accuracy.
Both counterfactual and synthetic data out-perform the \masklm{} objective. 
Our \emph{Large} model outperforms the \masklm{} model by almost 4 points on both metrics.
Our best \emph{Base} model is comparable to the previous \sota{}.
\errordescr.
}
\label{tab:sqa_results}
\end{table}

\paragraph{Efficiency} 

As discussed in Section \ref{sec:pruning} and Appendix \ref{sec:apx-train-time}, we can increase the model efficiency by reducing the input length.
By pruning the input of the \tabfact{} data we can improve training as well as inference time.
We compare pruning with the heuristic entity linking (HEL) \cite{2019TabFactA} and heuristic exact match (HEM) to different target lengths. 
We also studied other pruning methods, the results are reported in Appendix \ref{sec:apx-table-pruning}.
In Table \ref{tab:result-pruning} we find that HEM consistently outperforms HEL.
The best model at length $256$, while twice as fast to train (and apply), is only $0.8$ points behind the best full length model.
Even the model with length $128$, while using a much shorter length, out-performs \tablebert{} by more than $7$ points.

Given a pre-trained \masklm{} model our training consists of training on the artificial pre-training data and then fine-tuning on \tabfact{}.
We can therefore improve the training time by pre-training with shorter input sizes.
Table \ref{tab:result-pruning} shows that $512$ and $256$ give similar accuracy while the results for $128$ are about $1$ point lower.

\begin{table}[!t]
\centering
\pruningtable{short}
\label{tab:result-pruning}
\end{table}
\footnotetext{Not explicitly mentioned in the paper but implied by the batch size given (6) and the defaults in the code.}

\section{Analysis}
\label{sec:analysis}

\paragraph{Salient Groups}

\begin{figure*}[!t]
\small
\centering
\begin{tabular}{l|p{6cm}p{6cm}}
\textbf{Group} &	\textbf{Consistently Better} & \textbf{Persisting Errors} \\
\hline
\textbf{Aggregations} & 
Choi Moon - Sik played in Seoul three times in total. &
The total number of bronze medals were half of the total number of medals.
\\
\textbf{Superlatives} & 
Mapiu school has the highest roll in the state authority. &
Carlos Moya won the most tournaments with two wins.
\\
\textbf{Comparatives} & 
Bernard Holsey has 3 more yards than Angel Rubio. &
In 1982, the Kansas City Chiefs played more away games than home games.
\\
\textbf{Negations}    & 
The Warriors were not the home team at the game on 11-24-2006. &
Dean Semmens is not one of the four players born after 1981.
\end{tabular}%
\caption{
On the left column we show examples that our model gets correct for most runs and that \masklm gets wrong for most runs. The right column shows examples that the model continues to make mistakes on. Many of those include deeper chains of reasoning or more complex numeric operations.}
\label{fig:errors}
\vspace{-3ex}
\end{figure*}

\begin{table}[!t]
\small
\centering
\resizebox{1.0\columnwidth}{!}{
\begin{tabular}{lr|rr|rrr} \toprule			
    &               &  \multicolumn{2}{c|}{\textbf{C+S}} & \multicolumn{3}{c}{\textbf{\masklm}} \\
	& \textbf{Size}	&	\textbf{Acc}	&	\textbf{ER}	&	\textbf{Acc}	&	\bm{$\Delta_{\mbox{\textbf{Acc}}}$}	&	\bm{$\Delta_{\mbox{\textbf{ER}}}$}	\\
\midrule	
\textbf{Validation}	& $	100.0	$ & $	78.6	    $ & $  21.4 $ & $	69.6	$ & $	9.0	    $ &	 $  9.0    $	\\
\midrule
\textbf{Superlatives}  & $13.4$  & $79.6$  & $2.7$   & $66.9$  & $12.6$  & $1.7$   \\
\textbf{Aggregations}  & $11.6$  & $71.1$  & $3.4$   & $62.3$  & $8.9$   & $1.0$   \\
\textbf{Comparatives}  & $10.4$  & $72.3$  & $2.9$   & $62.6$  & $9.7$   & $1.0$   \\
\textbf{Negations}     & $3.3$   & $72.6$  & $0.9$   & $60.5$  & $12.1$  & $0.4$   \\
\midrule
\textbf{Multiple of the above}  & $9.2$   & $72.0$  & $2.6$   & $63.9$  & $8.2$   & $0.8$   \\
\textbf{Other}         & $51.9$  & $82.6$  & $9.1$   & $75.2$  & $7.4$   & $3.8$   \\
\bottomrule \end{tabular}
}
\caption{Comparing accuracy and total error rate (ER) for counterfactual and synthetic (C+S) and \masklm. 
Groups are derived from word heuristics.
The error rate in each group is taken with respect to the full set.
Negations and superlatives show the highest relative gains.
}
\vspace{-3ex}
\label{tab:slices}
\end{table}

To obtain detailed information of the improvements of our approach, we manually annotated $200$ random examples with the complex operations needed to answer them. 
We found $4$ salient groups: \textbf{Aggregations}, \textbf{superlatives}, \textbf{comparatives} and \textbf{negations}, and sort pairs into these groups via keywords in the text. To make the groups exclusive, we add a fifth case when more than one operation is needed. The accuracy of the heuristics was validated through further manual inspection of $50$ samples per group. The trigger words of each group are described in Appendix \ref{sec:slice_words}.

For each group within the validation set, we look at the difference in accuracy between different models. We also look at how the total error rate is divided among the groups as a way to guide the focus on pre-training tasks and modeling. The error rate defined in this way measures potential accuracy gains if all the errors in a group $S$ were fixed: $\text{ER}(S) = \frac{|\{ \text{ Errors in } S\}|}{|\{\text{ Validation examples}\}|}$.

Among the groups, the intermediate task data improve \textbf{superlatives} ($39\%$ error reduction) and \textbf{negations} ($31\%$) most (Table \ref{tab:slices}).
For example, we see that the accuracy is higher for \textbf{superlatives} than the for the overall validation set.

In Figure~\ref{fig:errors} we show examples in every group where our model is correct on the majority of the cases (across $9$ trials), and the \masklm baseline is not. We also show examples that continue to produce errors after our pre-training. Many examples in this last group require multi-hop reasoning or complex numerical operations.

\paragraph{Model Agreement}

\begin{figure}[!t]
\includegraphics[width=1\linewidth]{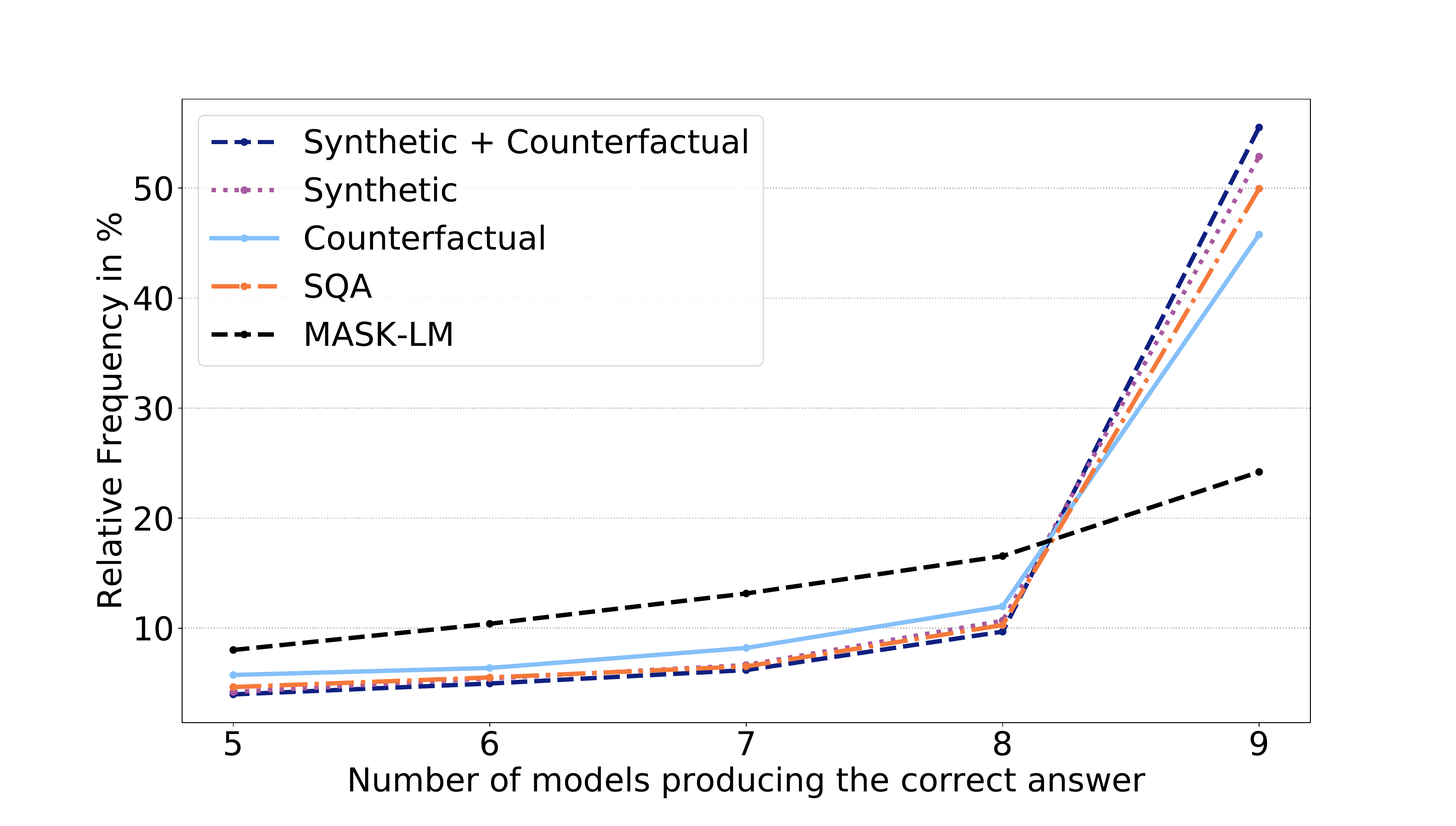}
\caption{
Frequency of the number of models that give the correct answer, out of 9 runs.
Better pre-training leads to more consistency across models.
The ratio of samples answered correctly by all models is $24.2\%$ for \masklm but $55.5\%$ for Synthetic + Counterfactual.
}
\label{fig:agreement}
\end{figure}

Similar to other complex binary classification datasets such as \boolq{}~\cite{clark-etal-2019-boolq}, for \tabfact{} one may question whether models are guessing the right answer.
To detect the magnitude of this issue we look at $9$ independent runs of each variant and analyze how many of them agree on the correct answer. 
Figure \ref{fig:agreement} shows that while for \masklm only for $24.2\%$ of the examples all models agree on the right answer, it goes up to $55.5\%$ when using using the counterfactual and synthetic pre-training. This suggests that the amount of guessing decreases substantially.

\section{Related Work}
\label{sec:related}

\paragraph{Logic-free Semantic Parsing}
Recently, methods that skip creating logical forms and generate answers directly have been used successfully for semantic parsing \cite{mueller-2019}.
In this group, \tapas{}~\cite{herzig-2020} uses special learned embeddings to encode row/column index and numerical order and pretrains a \masklm model on a large corpus of text and tables co-occurring on Wikipedia articles. 
Importantly, \emph{next sentence prediction} 
from \citet{devlin-19}, 
which in this context amounts to detecting whether the table and the sentence appear in the same article, was not found to be effective. Our hypothesis is that the task was not hard enough to provide a training signal.
We build on top of the \tapas~model and propose harder and more effective pre-training tasks to achieve strong performance on the \tabfact~dataset.

\paragraph{Entailment tasks}
Recognizing entailment has a long history in NLP~\cite{dagan-10}. Recently, the text to text framework has been expanded to incorporate structured data, like knowledge graphs \cite{vlachos-riedel-2015-identification}, tables \cite{jo2019aggchecker,gupta2020infotabs} or images \cite{suhr2017corpus,suhr2019corpus}.
The large-scale \tabfact~dataset~\cite{2019TabFactA} is one such example. Among the top performing models in the task is a BERT based model, acting on a flattened versioned of the table using textual templates to make the tables resemble natural text. Our approach has two key improvements: the usage of special embeddings, as introduced in \citet{herzig-2020}, and our novel \emph{counterfactual and synthetic pre-training} (Section~\ref{sec:methods}).

\paragraph{Pre-training objectives}
\emph{Next Sentence Prediction} (NSP) was introduced in \citet{devlin-19}, but follow-up work such as \citet{liu2019} identified that it did not contribute to model performance in some tasks. 
Other studies have found that application specific self-supervised pre-training objectives can improve performance of \masklm~models. One examples of such an objective is the \emph{Inverse Cloze Task} (ICT) \cite{lee-19}, that uses in-batch negatives and a two-tower dot-product similarity metric. 
\citet{Chang2020Pre-training} further expands on this idea and uses hyperlinks in Wikipedia as a weak label for topic overlap.

\paragraph{Intermediate Pre-training}
Language model fine-tuning \cite{howard-2018-universal} also know as domain adaptive pre-training \cite{gururangan-2020-dont-stop} has been studied as a way to handle covariate shift.
Our work is closer to intermediate task fine-tuning \cite{pruksachatkun-2020-intermediate-task} 
where one tries to teach the model \emph{higher-level abilities}.
Similarly we try to improve the discrete and numeric reasoning capabilities of the model.

\paragraph{Counterfactual data generation}
The most similar approach to ours appears in \citet{Xiong2020Pretrained}, replacing entities in Wikipedia by others with the same type for a \masklm~model objective. We, on the one hand, take advantage of other rows in the table to produce plausible negatives, and also replace dates and numbers.
Recently, \citet{Kaushik2020Learning,gardner-2020} have shown that exposing models to pairs of examples which are similar but have different labels can help to improve generalization, in some sense our \emph{Counterfactual} task is a heuristic version of this, that does not rely on manual annotation. \citet{sellam-2020-BLEURT} use perturbations of Wikipedia sentences for intermediate pre-training of a learned metric for text generation.

\paragraph{Numeric reasoning}
Numeric reasoning in Natural Language processing has been recognized as an
important part in entailment models \cite{sammons2010ask} and reading comprehension \cite{ran2019numnet}.
\citet{wallace-etal-2019-nlp} studied the capacity of different models on understanding numerical operations and show that \bert{}-based model still have headroom. This motivates the use of the synthetic generation approach to improve numerical reasoning in our model. 

\paragraph{Synthetic data generation} Synthetic data has been used to improve learning in \abr{nlp} tasks \cite{alberti-etal-2019-synthetic,lewis-etal-2019-unsupervised,wu-etal-2016-bilingually, leonandya-etal-2019-fast}. In semantic parsing for example \cite{wang-etal-2015-building, iyer-17, dbpal}, templates are used to bootstrap models that map text to logical forms or SQL.
\citet{salvatore-etal-2019-logical} use synthetic data generated from logical forms to evaluate the performance of textual entailment models (e.g., \bert{}).
\citet{geiger-2019-posing} use synthetic data to create \emph{fair} evaluation sets for natural language inference.
\citet{Geva2020InjectingNR} show the importance of injecting numerical reasoning via generated data into the model to solve reading comprehension tasks. They propose different templates for generating synthetic numerical examples. In our work we use a method that is better suited for tables and to the entailment task, and is arguably simpler.

\section{Conclusion}
\label{sec:conclusion}

We introduced two pre-training tasks, counterfactual and synthetic, to obtain \sota results on the \tabfact\cite{2019TabFactA} entailment task on tabular data.
We adapted the \bert-based architecture of \tapas\cite{herzig-2020} to binary classification and showed that pre-training on both tasks yields substantial improvements on \tabfact but also on a QA dataset, \sqa~\cite{iyyer-etal-2017-search}, even with only a subset of the training data. 

We ran a study on column selection methods to speed-up training and inference.
We found that we can speed up the model by a factor of 2 at a moderate drop in accuracy ($\approx 1$ point)
and by a factor of 4 at a larger drop but still with higher accuracy than previous approaches.

We characterized the complex operations required for table entailment to guide future research in this topic.
Our code and models will be open-sourced.

\section*{Acknowledgments}
We would like to thank Jordan Boyd-Graber, Yasemin Altun, Emily Pitler, Benjamin Boerschinger, William Cohen, Jonathan Herzig, Slav Petrov, and
the anonymous reviewers for their time, constructive
feedback, useful comments and suggestions about this work. 

\bibliographystyle{style/acl_natbib}
\bibliography{bib/journal-full,bib/jbg}

\begin{thebibliography}{52}
\expandafter\ifx\csname natexlab\endcsname\relax\def\natexlab#1{#1}\fi

\bibitem[{Alberti et~al.(2019)Alberti, Andor, Pitler, Devlin, and
  Collins}]{alberti-etal-2019-synthetic}
Chris Alberti, Daniel Andor, Emily Pitler, Jacob Devlin, and Michael Collins.
  2019.
\newblock \href {https://doi.org/10.18653/v1/P19-1620} {Synthetic {QA} corpora
  generation with roundtrip consistency}.
\newblock In \emph{Proceedings of the Association for Computational
  Linguistics}, pages 6168--6173, Florence, Italy. Association for
  Computational Linguistics.

\bibitem[{Bowman et~al.(2015)Bowman, Angeli, Potts, and
  Manning}]{bowman-etal-2015-large}
Samuel~R. Bowman, Gabor Angeli, Christopher Potts, and Christopher~D. Manning.
  2015.
\newblock \href {https://doi.org/10.18653/v1/D15-1075} {A large annotated
  corpus for learning natural language inference}.
\newblock In \emph{Proceedings of Empirical Methods in Natural Language
  Processing}, pages 632--642, Lisbon, Portugal. Association for Computational
  Linguistics.

\bibitem[{Chang et~al.(2020)Chang, Yu, Chang, Yang, and
  Kumar}]{Chang2020Pre-training}
Wei-Cheng Chang, Felix~X. Yu, Yin-Wen Chang, Yiming Yang, and Sanjiv Kumar.
  2020.
\newblock \href {https://openreview.net/forum?id=rkg-mA4FDr} {Pre-training
  tasks for embedding-based large-scale retrieval}.
\newblock In \emph{Proceedings of the International Conference on Learning
  Representations}, Addis Ababa, Ethiopia.

\bibitem[{Chen et~al.(2020)Chen, Wang, Chen, Zhang, Wang, Li, Zhou, and
  Wang}]{2019TabFactA}
Wenhu Chen, Hongmin Wang, Jianshu Chen, Yunkai Zhang, Hong Wang, Shiyang Li,
  Xiyou Zhou, and William~Yang Wang. 2020.
\newblock \href {https://openreview.net/forum?id=rkeJRhNYDH} {Tabfact: A
  large-scale dataset for table-based fact verification}.
\newblock In \emph{Proceedings of the International Conference on Learning
  Representations}, Addis Ababa, Ethiopia.

\bibitem[{Clark et~al.(2019)Clark, Lee, Chang, Kwiatkowski, Collins, and
  Toutanova}]{clark-etal-2019-boolq}
Christopher Clark, Kenton Lee, Ming-Wei Chang, Tom Kwiatkowski, Michael
  Collins, and Kristina Toutanova. 2019.
\newblock \href {https://doi.org/10.18653/v1/N19-1300} {{B}ool{Q}: Exploring
  the surprising difficulty of natural yes/no questions}.
\newblock In \emph{Conference of the North American Chapter of the Association
  for Computational Linguistics}, pages 2924--2936, Minneapolis, Minnesota.
  Association for Computational Linguistics.

\bibitem[{Clark et~al.(2020)Clark, Luong, Le, and Manning}]{Clark2020ELECTRA:}
Kevin Clark, Minh-Thang Luong, Quoc~V. Le, and Christopher~D. Manning. 2020.
\newblock \href {https://openreview.net/forum?id=r1xMH1BtvB} {Electra:
  Pre-training text encoders as discriminators rather than generators}.
\newblock In \emph{Proceedings of the International Conference on Learning
  Representations}, Addis Ababa, Ethiopia.

\bibitem[{Clark(2019)}]{aristo}
Peter Clark. 2019.
\newblock \href {https://doi.org/10.1145/3360901.3364451} {Project aristo:
  Towards machines that capture and reason with science knowledge}.
\newblock In \emph{Proceedings of the 10th International Conference on
  Knowledge Capture}, K-CAP ’19, page 1–2, New York, NY, USA. Association
  for Computing Machinery.

\bibitem[{Conneau et~al.(2017)Conneau, Kiela, Schwenk, Barrault, and
  Bordes}]{conneau-etal-2017-supervised}
Alexis Conneau, Douwe Kiela, Holger Schwenk, Lo{\"\i}c Barrault, and Antoine
  Bordes. 2017.
\newblock \href {https://doi.org/10.18653/v1/D17-1070} {Supervised learning of
  universal sentence representations from natural language inference data}.
\newblock In \emph{Proceedings of Empirical Methods in Natural Language
  Processing}, pages 670--680, Copenhagen, Denmark. Association for
  Computational Linguistics.

\bibitem[{Dagan et~al.(2010)Dagan, Dolan, Magnini, and Roth}]{dagan-10}
Ido Dagan, Bill Dolan, Bernardo Magnini, and Dan Roth. 2010.
\newblock Recognizing textual entailment: Rationale, evaluation and approaches.
\newblock \emph{Journal of Natural Language Engineering}, 4.

\bibitem[{Dagan et~al.(2005)Dagan, Glickman, and Magnini}]{dagan2005pascal}
Ido Dagan, Oren Glickman, and Bernardo Magnini. 2005.
\newblock The pascal recognising textual entailment challenge.
\newblock In \emph{Machine Learning Challenges Workshop}, pages 177--190.
  Springer.

\bibitem[{Devlin et~al.(2019)Devlin, Chang, Lee, and Toutanova}]{devlin-19}
Jacob Devlin, Ming-Wei Chang, Kenton Lee, and Kristina Toutanova. 2019.
\newblock \href {https://doi.org/10.18653/v1/N19-1423} {{BERT}: Pre-training of
  deep bidirectional transformers for language understanding}.
\newblock In \emph{Conference of the North American Chapter of the Association
  for Computational Linguistics}, pages 4171--4186, Minneapolis, Minnesota.
  Association for Computational Linguistics.

\bibitem[{Dodge et~al.(2019)Dodge, Gururangan, Card, Schwartz, and
  Smith}]{dodge-19}
Jesse Dodge, Suchin Gururangan, Dallas Card, Roy Schwartz, and Noah~A. Smith.
  2019.
\newblock \href {https://doi.org/10.18653/v1/D19-1224} {Show your work:
  Improved reporting of experimental results}.
\newblock In \emph{Proceedings of Empirical Methods in Natural Language
  Processing}, pages 2185--2194, Hong Kong, China. Association for
  Computational Linguistics.

\bibitem[{Gardner et~al.(2020)Gardner, Artzi, Basmova, Berant, Bogin, Chen,
  Dasigi, Dua, Elazar, Gottumukkala, Gupta, Hajishirzi, Ilharco, Khashabi, Lin,
  Liu, Liu, Mulcaire, Ning, Singh, Smith, Subramanian, Tsarfaty, Wallace,
  Zhang, and Zhou}]{gardner-2020}
Matt Gardner, Yoav Artzi, Victoria Basmova, Jonathan Berant, Ben Bogin, Sihao
  Chen, Pradeep Dasigi, Dheeru Dua, Yanai Elazar, Ananth Gottumukkala, Nitish
  Gupta, Hanna Hajishirzi, Gabriel Ilharco, Daniel Khashabi, Kevin Lin,
  Jiangming Liu, Nelson~F. Liu, Phoebe Mulcaire, Qiang Ning, Sameer Singh,
  Noah~A. Smith, Sanjay Subramanian, Reut Tsarfaty, Eric Wallace, Ally Zhang,
  and Ben Zhou. 2020.
\newblock \href {http://arxiv.org/abs/2004.02709} {Evaluating {NLP} models via
  contrast sets}.
\newblock \emph{CoRR}, abs/2004.02709.

\bibitem[{Geiger et~al.(2019)Geiger, Cases, Karttunen, and
  Potts}]{geiger-2019-posing}
Atticus Geiger, Ignacio Cases, Lauri Karttunen, and Christopher Potts. 2019.
\newblock \href {https://doi.org/10.18653/v1/D19-1456} {Posing fair
  generalization tasks for natural language inference}.
\newblock In \emph{Proceedings of Empirical Methods in Natural Language
  Processing}, pages 4485--4495, Hong Kong, China. Association for
  Computational Linguistics.

\bibitem[{Geva et~al.(2020)Geva, Gupta, and Berant}]{Geva2020InjectingNR}
Mor Geva, Ankit Gupta, and Jonathan Berant. 2020.
\newblock \href {https://doi.org/10.18653/v1/2020.acl-main.89} {Injecting
  numerical reasoning skills into language models}.
\newblock In \emph{acl}, pages 946--958, Online. Association for Computational
  Linguistics.

\bibitem[{Golovin et~al.(2017)Golovin, Solnik, Moitra, Kochanski, Karro, and
  Sculley}]{vizier}
Daniel Golovin, Benjamin Solnik, Subhodeep Moitra, Greg Kochanski, John~Elliot
  Karro, and D.~Sculley, editors. 2017.
\newblock \href
  {http://www.kdd.org/kdd2017/papers/view/google-vizier-a-service-for-black-box-optimization}
  {\emph{Google Vizier: A Service for Black-Box Optimization}}.

\bibitem[{Gupta et~al.(2020)Gupta, Mehta, Nokhiz, and
  Srikumar}]{gupta2020infotabs}
Vivek Gupta, Maitrey Mehta, Pegah Nokhiz, and Vivek Srikumar. 2020.
\newblock \href {https://arxiv.org/abs/2005.06117} {Infotabs: Inference on
  tables as semi-structured data}.
\newblock In \emph{Proceedings of the Association for Computational
  Linguistics}, Seattle, Washington. Association for Computational Linguistics.

\bibitem[{Gururangan et~al.(2020)Gururangan, Marasovi{\'c}, Swayamdipta, Lo,
  Beltagy, Downey, and Smith}]{gururangan-2020-dont-stop}
Suchin Gururangan, Ana Marasovi{\'c}, Swabha Swayamdipta, Kyle Lo, Iz~Beltagy,
  Doug Downey, and Noah~A. Smith. 2020.
\newblock \href {https://arxiv.org/abs/2004.10964} {Don't stop pretraining:
  Adapt language models to domains and tasks}.
\newblock In \emph{Proceedings of the Association for Computational
  Linguistics}, Seattle, Washington. Association for Computational Linguistics.

\bibitem[{Henderson et~al.(2018)Henderson, Islam, Bachman, Pineau, Precup, and
  Meger}]{AAAI1816669}
Peter Henderson, Riashat Islam, Philip Bachman, Joelle Pineau, Doina Precup,
  and David Meger. 2018.
\newblock \href
  {https://www.aaai.org/ocs/index.php/AAAI/AAAI18/paper/view/16669} {Deep
  reinforcement learning that matters}.
\newblock In \emph{Association for the Advancement of Artificial Intelligence}.

\bibitem[{Herzig et~al.(2020)Herzig, Nowak, M{\"u}ller, Piccinno, and
  Eisenschlos}]{herzig-2020}
Jonathan Herzig, Pawel~Krzysztof Nowak, Thomas M{\"u}ller, Francesco Piccinno,
  and Julian Eisenschlos. 2020.
\newblock \href {https://doi.org/10.18653/v1/2020.acl-main.398} {{T}a{P}as:
  Weakly supervised table parsing via pre-training}.
\newblock In \emph{Proceedings of the Association for Computational
  Linguistics}, pages 4320--4333, Online. Association for Computational
  Linguistics.

\bibitem[{Howard and Ruder(2018)}]{howard-2018-universal}
Jeremy Howard and Sebastian Ruder. 2018.
\newblock \href {https://doi.org/10.18653/v1/P18-1031} {Universal language
  model fine-tuning for text classification}.
\newblock In \emph{Proceedings of the Association for Computational
  Linguistics}, pages 328--339, Melbourne, Australia. Association for
  Computational Linguistics.

\bibitem[{Iyer et~al.(2017)Iyer, Dandekar, , and Csernai}]{iyer-17}
Shankar Iyer, Nikhil Dandekar, , and Korn\'el Csernai. 2017.
\newblock \href
  {https://www.quora.com/q/quoradata/First-Quora-Dataset-Release-Question-Pairs}
  {Quora question pairs}.

\bibitem[{Iyyer et~al.(2017)Iyyer, Yih, and Chang}]{iyyer-etal-2017-search}
Mohit Iyyer, Wen-tau Yih, and Ming-Wei Chang. 2017.
\newblock \href {https://doi.org/10.18653/v1/P17-1167} {Search-based neural
  structured learning for sequential question answering}.
\newblock In \emph{Proceedings of the Association for Computational
  Linguistics}, pages 1821--1831, Vancouver, Canada. Association for
  Computational Linguistics.

\bibitem[{Jo et~al.(2019)Jo, Trummer, Yu, Wang, Yu, Liu, and
  Mehta}]{jo2019aggchecker}
Saehan Jo, Immanuel Trummer, Weicheng Yu, Xuezhi Wang, Cong Yu, Daniel Liu, and
  Niyati Mehta. 2019.
\newblock \href {https://doi.org/10.14778/3352063.3352104} {Aggchecker: A
  fact-checking system for text summaries of relational data sets}.
\newblock \emph{International Conference on Very Large Databases},
  12(12):1938–1941.

\bibitem[{Kaushik et~al.(2020)Kaushik, Hovy, and Lipton}]{Kaushik2020Learning}
Divyansh Kaushik, Eduard~H. Hovy, and Zachary~Chase Lipton. 2020.
\newblock \href {https://openreview.net/forum?id=Sklgs0NFvr} {Learning the
  difference that makes {A} difference with counterfactually-augmented data}.
\newblock In \emph{Proceedings of the International Conference on Learning
  Representations}, Addis Ababa, Ethiopia.

\bibitem[{Khashabi et~al.(2016)Khashabi, Khot, Sabharwal, Clark, Etzioni, and
  Roth}]{Khashabi19}
Daniel Khashabi, Tushar Khot, Ashish Sabharwal, Peter Clark, Oren Etzioni, and
  Dan Roth. 2016.
\newblock \href {https://www.ijcai.org/Abstract/16/166} {Question answering via
  integer programming over semi-structured knowledge}.
\newblock In \emph{International Joint Conference on Artificial Intelligence},
  IJCAI’16, page 1145–1152. AAAI Press.

\bibitem[{Kiss and Strunk(2006)}]{kiss-06}
Tibor Kiss and Jan Strunk. 2006.
\newblock \href {https://doi.org/10.1162/coli.2006.32.4.485} {Unsupervised
  multilingual sentence boundary detection}.
\newblock \emph{Computational Linguistics}, 32(4):485--525.

\bibitem[{Lee et~al.(2019)Lee, Chang, and Toutanova}]{lee-19}
Kenton Lee, Ming-Wei Chang, and Kristina Toutanova. 2019.
\newblock \href {https://doi.org/10.18653/v1/P19-1612} {Latent retrieval for
  weakly supervised open domain question answering}.
\newblock In \emph{Proceedings of the Association for Computational
  Linguistics}, pages 6086--6096, Florence, Italy. Association for
  Computational Linguistics.

\bibitem[{Leonandya et~al.(2019)Leonandya, Hupkes, Bruni, and
  Kruszewski}]{leonandya-etal-2019-fast}
Rezka Leonandya, Dieuwke Hupkes, Elia Bruni, and Germ{\'a}n Kruszewski. 2019.
\newblock \href {https://doi.org/10.18653/v1/W19-0419} {The fast and the
  flexible: Training neural networks to learn to follow instructions from small
  data}.
\newblock In \emph{Proceedings of the International Conference on Computational
  Semantics}, pages 223--234, Gothenburg, Sweden. Association for Computational
  Linguistics.

\bibitem[{Lewis et~al.(2019)Lewis, Denoyer, and
  Riedel}]{lewis-etal-2019-unsupervised}
Patrick Lewis, Ludovic Denoyer, and Sebastian Riedel. 2019.
\newblock \href {https://doi.org/10.18653/v1/P19-1484} {Unsupervised question
  answering by cloze translation}.
\newblock In \emph{Proceedings of the Association for Computational
  Linguistics}, pages 4896--4910, Florence, Italy. Association for
  Computational Linguistics.

\bibitem[{Liu et~al.(2019)Liu, Ott, Goyal, Du, Joshi, Chen, Levy, Lewis,
  Zettlemoyer, and Stoyanov}]{liu2019}
Yinhan Liu, Myle Ott, Naman Goyal, Jingfei Du, Mandar Joshi, Danqi Chen, Omer
  Levy, Mike Lewis, Luke Zettlemoyer, and Veselin Stoyanov. 2019.
\newblock \href {http://arxiv.org/abs/1907.11692} {Roberta: {A} robustly
  optimized {BERT} pretraining approach}.
\newblock \emph{CoRR}, abs/1907.11692.

\bibitem[{Loper and Bird(2002)}]{loper-02}
Edward Loper and Steven Bird. 2002.
\newblock {NLTK}: the natural language toolkit.
\newblock In \emph{Tools and methodologies for teaching}.

\bibitem[{Mikolov et~al.(2013)Mikolov, Sutskever, Chen, Corrado, and
  Dean}]{mikolov-13}
Tomas Mikolov, Ilya Sutskever, Kai Chen, Greg Corrado, and Jeffrey Dean. 2013.
\newblock \href
  {http://papers.nips.cc/paper/5021-distributed-representations-of-words-and-phrases-and-their-compositionality}
  {Distributed representations of words and phrases and their
  compositionality}.
\newblock In \emph{Proceedings of Advances in Neural Information Processing
  Systems}, NIPS’13, page 3111–3119, Lake Tahoe, Nevada. Curran Associates
  Inc.

\bibitem[{Mueller et~al.(2019)Mueller, Piccinno, Shaw, Nicosia, and
  Altun}]{mueller-2019}
Thomas Mueller, Francesco Piccinno, Peter Shaw, Massimo Nicosia, and Yasemin
  Altun. 2019.
\newblock \href {https://doi.org/10.18653/v1/D19-1603} {Answering
  conversational questions on structured data without logical forms}.
\newblock In \emph{Proceedings of Empirical Methods in Natural Language
  Processing}, pages 5902--5910, Hong Kong, China. Association for
  Computational Linguistics.

\bibitem[{Pasupat and Liang(2015)}]{pasupat2015compositional}
Panupong Pasupat and Percy Liang. 2015.
\newblock \href {https://doi.org/10.3115/v1/P15-1142} {Compositional semantic
  parsing on semi-structured tables}.
\newblock In \emph{Proceedings of the Association for Computational
  Linguistics}, pages 1470--1480, Beijing, China. Association for Computational
  Linguistics.

\bibitem[{Pruksachatkun et~al.(2020)Pruksachatkun, Phang, Liu, Htut, Zhang,
  Pang, Vania, Kann, and Bowman}]{pruksachatkun-2020-intermediate-task}
Yada Pruksachatkun, Jason Phang, Haokun Liu, Phu~Mon Htut, Xiaoyi Zhang,
  Richard~Yuanzhe Pang, Clara Vania, Katharina Kann, and Samuel~R. Bowman.
  2020.
\newblock \href {https://arxiv.org/abs/2005.00628} {Intermediate-task transfer
  learning with pretrained models for natural language understanding: When and
  why does it work?}
\newblock In \emph{Proceedings of the Association for Computational
  Linguistics}, Seattle, Washington. Association for Computational Linguistics.

\bibitem[{Ran et~al.(2019)Ran, Lin, Li, Zhou, and Liu}]{ran2019numnet}
Qiu Ran, Yankai Lin, Peng Li, Jie Zhou, and Zhiyuan Liu. 2019.
\newblock \href {https://doi.org/10.18653/v1/D19-1251} {{N}um{N}et: Machine
  reading comprehension with numerical reasoning}.
\newblock In \emph{Proceedings of Empirical Methods in Natural Language
  Processing}, pages 2474--2484, Hong Kong, China. Association for
  Computational Linguistics.

\bibitem[{Salvatore et~al.(2019)Salvatore, Finger, and
  Hirata~Jr}]{salvatore-etal-2019-logical}
Felipe Salvatore, Marcelo Finger, and Roberto Hirata~Jr. 2019.
\newblock \href {https://doi.org/10.18653/v1/D19-6103} {A logical-based corpus
  for cross-lingual evaluation}.
\newblock In \emph{Proceedings of the 2nd Workshop on Deep Learning Approaches
  for Low-Resource NLP (DeepLo 2019)}, pages 22--30, Hong Kong, China.
  Association for Computational Linguistics.

\bibitem[{Sammons et~al.(2010)Sammons, Vydiswaran, and Roth}]{sammons2010ask}
Mark Sammons, V.G.Vinod Vydiswaran, and Dan Roth. 2010.
\newblock \href {https://www.aclweb.org/anthology/P10-1122} {{``}ask not what
  textual entailment can do for you...{''}}.
\newblock In \emph{Proceedings of the Association for Computational
  Linguistics}, pages 1199--1208, Uppsala, Sweden. Association for
  Computational Linguistics.

\bibitem[{Sellam et~al.(2020)Sellam, Das, and Parikh}]{sellam-2020-BLEURT}
Thibault Sellam, Dipanjan Das, and Ankur~P. Parikh. 2020.
\newblock \href {https://arxiv.org/abs/2004.04696} {Bleurt: Learning robust
  metrics for text generation}.
\newblock In \emph{Proceedings of the Association for Computational
  Linguistics}, Seattle, Washington. Association for Computational Linguistics.

\bibitem[{Socher et~al.(2013)Socher, Bauer, Manning, and
  Ng}]{socher-etal-2013-parsing}
Richard Socher, John Bauer, Christopher~D. Manning, and Andrew~Y. Ng. 2013.
\newblock \href {https://www.aclweb.org/anthology/P13-1045} {Parsing with
  compositional vector grammars}.
\newblock In \emph{Proceedings of the Association for Computational
  Linguistics}, pages 455--465, Sofia, Bulgaria. Association for Computational
  Linguistics.

\bibitem[{Subramanian et~al.(2018)Subramanian, Trischler, Bengio, and
  Pal}]{subramanian2018learning}
Sandeep Subramanian, Adam Trischler, Yoshua Bengio, and Christopher~J Pal.
  2018.
\newblock \href {https://openreview.net/forum?id=B18WgG-CZ} {Learning general
  purpose distributed sentence representations via large scale multi-task
  learning}.
\newblock In \emph{Proceedings of the International Conference on Learning
  Representations}.

\bibitem[{Suhr et~al.(2017)Suhr, Lewis, Yeh, and Artzi}]{suhr2017corpus}
Alane Suhr, Mike Lewis, James Yeh, and Yoav Artzi. 2017.
\newblock \href {https://doi.org/10.18653/v1/P17-2034} {A corpus of natural
  language for visual reasoning}.
\newblock In \emph{Proceedings of the Association for Computational
  Linguistics}, pages 217--223, Vancouver, Canada. Association for
  Computational Linguistics.

\bibitem[{Suhr et~al.(2019)Suhr, Zhou, Zhang, Zhang, Bai, and
  Artzi}]{suhr2019corpus}
Alane Suhr, Stephanie Zhou, Ally Zhang, Iris Zhang, Huajun Bai, and Yoav Artzi.
  2019.
\newblock \href {https://doi.org/10.18653/v1/P19-1644} {A corpus for reasoning
  about natural language grounded in photographs}.
\newblock In \emph{Proceedings of the Association for Computational
  Linguistics}, pages 6418--6428, Florence, Italy. Association for
  Computational Linguistics.

\bibitem[{Vaswani et~al.(2017)Vaswani, Shazeer, Parmar, Uszkoreit, Jones,
  Gomez, Kaiser, and Polosukhin}]{vaswani-2017}
Ashish Vaswani, Noam Shazeer, Niki Parmar, Jakob Uszkoreit, Llion Jones,
  Aidan~N Gomez, {\L}ukasz Kaiser, and Illia Polosukhin. 2017.
\newblock \href
  {http://papers.nips.cc/paper/7181-attention-is-all-you-need.pdf} {Attention
  is all you need}.
\newblock In I.~Guyon, U.~V. Luxburg, S.~Bengio, H.~Wallach, R.~Fergus,
  S.~Vishwanathan, and R.~Garnett, editors, \emph{Proceedings of Advances in
  Neural Information Processing Systems}, pages 5998--6008. Curran Associates,
  Inc.

\bibitem[{Vlachos and Riedel(2015)}]{vlachos-riedel-2015-identification}
Andreas Vlachos and Sebastian Riedel. 2015.
\newblock \href {https://doi.org/10.18653/v1/D15-1312} {Identification and
  verification of simple claims about statistical properties}.
\newblock In \emph{Proceedings of Empirical Methods in Natural Language
  Processing}, pages 2596--2601, Lisbon, Portugal. Association for
  Computational Linguistics.

\bibitem[{Wallace et~al.(2019)Wallace, Wang, Li, Singh, and
  Gardner}]{wallace-etal-2019-nlp}
Eric Wallace, Yizhong Wang, Sujian Li, Sameer Singh, and Matt Gardner. 2019.
\newblock \href {https://doi.org/10.18653/v1/D19-1534} {Do {NLP} models know
  numbers? probing numeracy in embeddings}.
\newblock In \emph{Proceedings of Empirical Methods in Natural Language
  Processing}, pages 5307--5315, Hong Kong, China. Association for
  Computational Linguistics.

\bibitem[{Wang et~al.(2015)Wang, Berant, and Liang}]{wang-etal-2015-building}
Yushi Wang, Jonathan Berant, and Percy Liang. 2015.
\newblock \href {https://doi.org/10.3115/v1/P15-1129} {Building a semantic
  parser overnight}.
\newblock In \emph{Proceedings of the Association for Computational
  Linguistics}, pages 1332--1342, Beijing, China. Association for Computational
  Linguistics.

\bibitem[{Weir et~al.(2020)Weir, Utama, Galakatos, Crotty, Ilkhechi, Ramaswamy,
  Bhushan, Geisler, H\"{a}ttasch, Eger, Cetintemel, and Binnig}]{dbpal}
Nathaniel Weir, Prasetya Utama, Alex Galakatos, Andrew Crotty, Amir Ilkhechi,
  Shekar Ramaswamy, Rohin Bhushan, Nadja Geisler, Benjamin H\"{a}ttasch,
  Steffen Eger, Ugur Cetintemel, and Carsten Binnig. 2020.
\newblock \href {https://doi.org/10.1145/3318464.3380589} {Dbpal: A fully
  pluggable nl2sql training pipeline}.
\newblock In \emph{Proceedings of the ACM SIGMOD International Conference on
  Management of Data}, SIGMOD ’20, page 2347–2361, New York, NY, USA.
  Association for Computing Machinery.

\bibitem[{Wu et~al.(2016)Wu, Shi, Chen, Huang, and
  Su}]{wu-etal-2016-bilingually}
Changxing Wu, Xiaodong Shi, Yidong Chen, Yanzhou Huang, and Jinsong Su. 2016.
\newblock \href {https://doi.org/10.18653/v1/D16-1253} {Bilingually-constrained
  synthetic data for implicit discourse relation recognition}.
\newblock In \emph{Proceedings of Empirical Methods in Natural Language
  Processing}, pages 2306--2312, Austin, Texas. Association for Computational
  Linguistics.

\bibitem[{Xiong et~al.(2020)Xiong, Du, Wang, and
  Stoyanov}]{Xiong2020Pretrained}
Wenhan Xiong, Jingfei Du, William~Yang Wang, and Veselin Stoyanov. 2020.
\newblock \href {https://openreview.net/forum?id=BJlzm64tDH} {Pretrained
  encyclopedia: Weakly supervised knowledge-pretrained language model}.
\newblock In \emph{Proceedings of the International Conference on Learning
  Representations}, Addis Ababa, Ethiopia.

\bibitem[{Zhong et~al.(2020)Zhong, Tang, Feng, Duan, Zhou, Gong, Shou, Jiang,
  Wang, and Yin}]{zhong2020logicalfactchecker}
Wanjun Zhong, Duyu Tang, Zhangyin Feng, Nan Duan, Ming Zhou, Ming Gong, Linjun
  Shou, Daxin Jiang, Jiahai Wang, and Jian Yin. 2020.
\newblock \href {https://arxiv.org/abs/2004.13659} {{LogicalFactChecker}:
  Leveraging logical operations for fact checking with graph module network}.
\newblock In \emph{Proceedings of the Association for Computational
  Linguistics}, Seattle, Washington. Association for Computational Linguistics.

\end{thebibliography}

\clearpage

\appendix\section*{Appendix}

We provide details on our experimental setup and hyper-parameter tuning in Section \ref{sec:apx-repro}.
Section \ref{sec:apx-input} and \ref{sec:apx-dataset-stats} give additional information on model and the \tabfact{}~dataset.
We give details and results regarding our column pruning approach in Section~\ref{sec:apx-table-pruning-descr}.
Full results for \sqa are displayed in Section ~\ref{sec:sqa_dev_results}. Section ~\ref{sec:apx-table-pruning} shows the accuracy on the pre-training tasks held-out sets.
Section \ref{sec:slice_words} contains the trigger words used for identifying the salient groups in the analysis section.

\section{Reproducibility}
\label{sec:apx-repro}

\subsection{Hyper-Parameter Search}

The hyper-parameters are optimized using a black box Bayesian optimizer similar to Google Vizier~\cite{vizier} which looked at validation accuracy after $8,000$ steps only, in order to prevent over-fitting and use resources effectively. 
The ranges used were a learning rate from $10^{-6}$ to $3\times 10^{-4}$, dropout probabilities from $0$ to $0.2$ and warm-up ratio from $0$ to $0.05$. We used $200$ runs and kept the median values for the top $20$ trials.

In order to show the impact of the number of trials in the expected validation results, we follow~\citet{AAAI1816669} and \citet{dodge-19}. Given that we used Bayesian optimization instead of random search, we applied the \emph{bootstrap} method to estimate mean and variance of the max validation accuracy at $8,000$ steps for different number of trials. From trial $10$ to $200$ we noted an increase  of $0.4\%$ in accuracy and a standard deviation that decreases from $2\%$ to $1.3\%$.

\subsection{Hyper-Parameters}

We use the same hyper-parameters for pre-training and fine-tuning.
For pre-training, the input length is 256 and 512 for fine-tuning if not stated otherwise.
We use $80,000$ training steps, a \textbf{learning rate of $\mathbf{2e^{-5}}$} and a \textbf{warm-up ratio of 0.05}.
We disable the attention dropout in \bert{} but use a \textbf{hidden dropout probability of 0.07 }.
Finally, we use an Adam optimizer with weight decay with the same configuration as \bert{}.

For \sqa we do not use any search algorithm and use the same model and the same hyper-parameters as the ones used in~\citet{herzig-2020}.
The only difference is that we start the fine-tuning from a checkpoint trained on our intermediate pre-training entailment task.

\subsection{Number of Parameters}

The number of parameters is the same as for \bert{}: $110M$ for base models and $340M$ for Large models.

\subsection{Training Time}
\label{sec:apx-train-time}

We train all our models on \textbf{Cloud TPUs V3}. The input length has a big impact on the processing speed of the batches and thus on the overall training time and training cost.
For a \bert{}-Base model during training, we can process approximately 8700 examples per second at input length 128, 5100 at input length 256 and 2600 at input length 512.
This corresponds to training times of approx. \textbf{78 minutes, 133 minutes and 262 minutes}, respectively.

A \bert{}-Large model  processes approximately 800 examples per second at length 512 and takes \textbf{14 hours} to train. 

\section{Model}
\label{sec:apx-input}

For illustrative purposes, we include the input representation using the $6$ types of embeddings, as depicted by \citet{herzig-2020}.

\begin{figure}
    \centering
    \scalebox{0.9}{
    \includegraphics[width=0.95\linewidth]{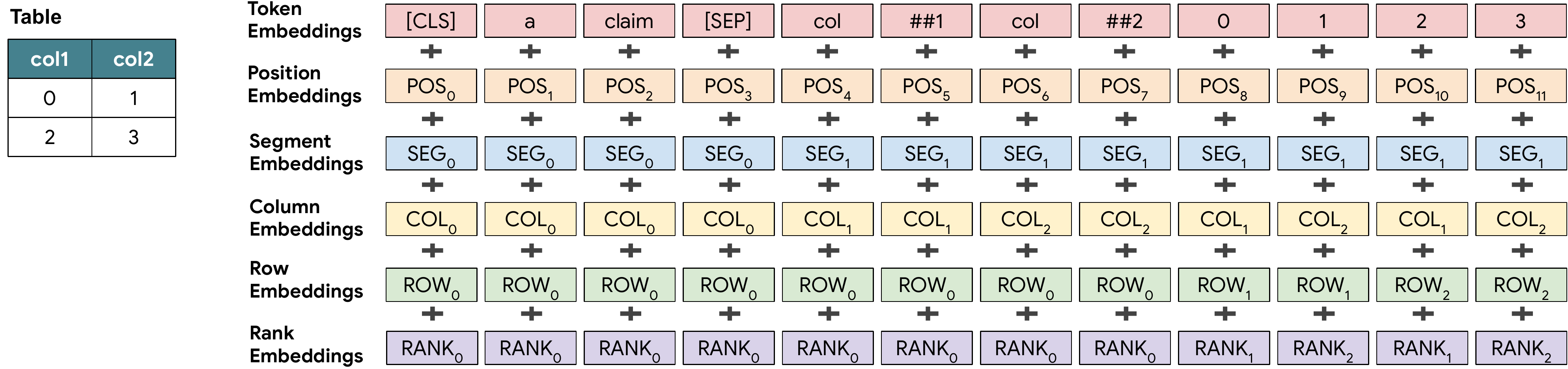}
    }
    \caption{Input representation for model.}
    \label{fig:apx-input}
\end{figure}

\section{Dataset}
\label{sec:apx-dataset-stats}

Statistics of the \tabfact dataset can be found in table \ref{tab:apx-dataset-stats}.

\begin{table}[H]
\begin{center}
\scalebox{0.9}{
\begin{tabular}{llll}
\toprule
   & \textbf{Statements} & \textbf{Tables} \\
\midrule
\textbf{Train}      & 92,283 & 13,182 \\
\textbf{Val}        & 12,792 & 1,696 \\
\textbf{Test}       & 12,779 & 1,695  \\
\midrule
\textbf{Total}      & 118,275 & 16,573 \\
\midrule
\textbf{Simple}     & 50,244 & 9,189 \\
\textbf{Complex}    & 68,031 & 7,392 \\
\bottomrule

\end{tabular}
}
\end{center}
\caption{\tabfact~ dataset statistics.}
\label{tab:apx-dataset-stats}
\end{table}

\begin{figure}[H]
    \scalebox{1.0}{
    \includegraphics[width=1\linewidth]{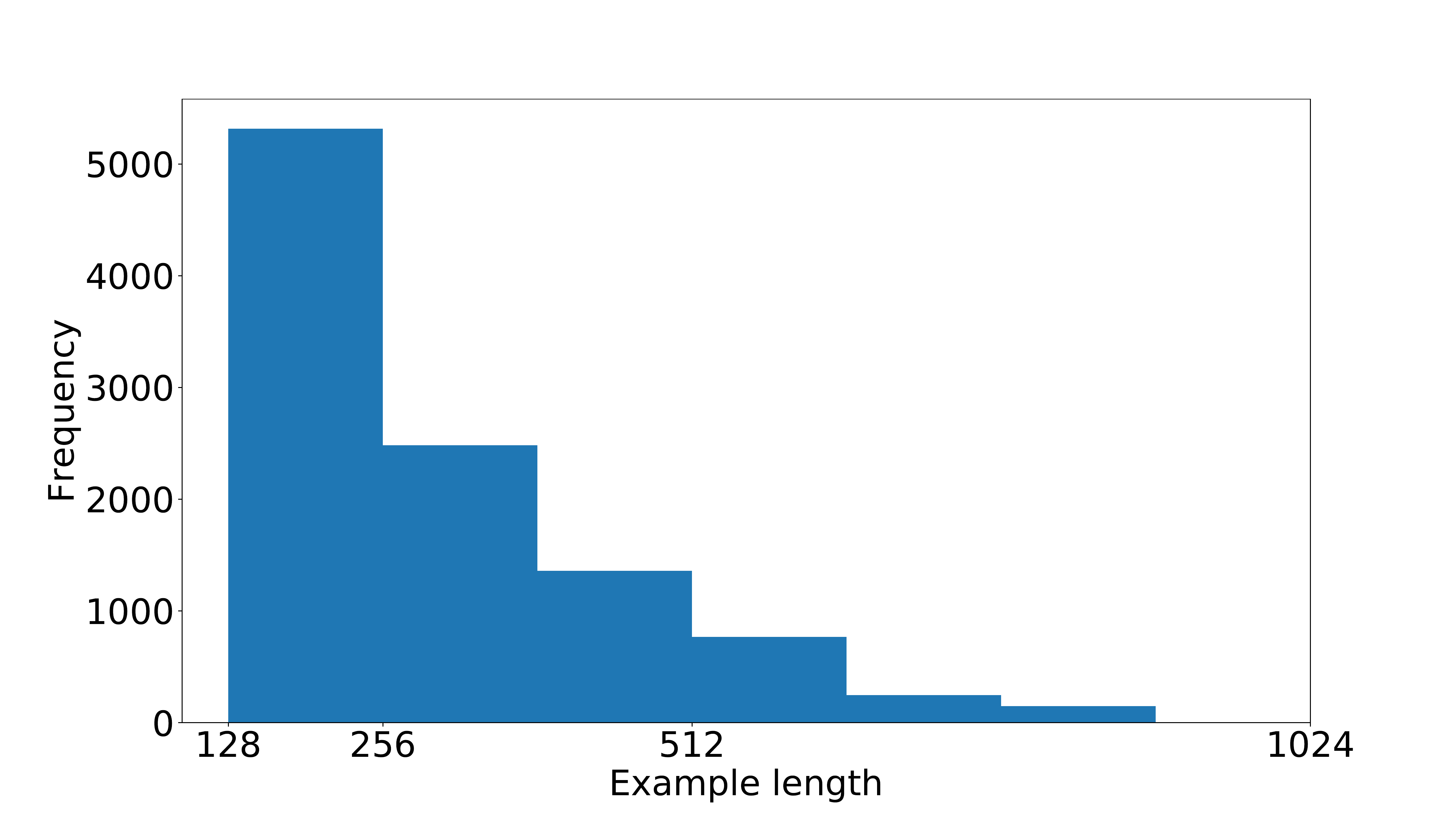}
    }
    \caption{Input length histogram for \tabfact{} validation dataset when tokenized with \bert{}~tokenizer.}
    \label{fig:examples_distribution}
\end{figure}

\section{Columns selection algorithm}
\label{sec:apx-table-pruning-descr}

Let $cost(.) \in \mathbb{N}$ be the function that computes the number of tokens given a text using the \bert{}~tokenizer, $t_s$ the tokenized statement text, $t_{c_i}$ the text of the column $i$. We denote the columns as $(c_1,..,c_n)$ ordered by their scores $$\forall i \in [1, ..n-1] f(c_i) > f(c_{i+1}) $$ where $n$ is the number of columns. Let $m$ be the maximum number of tokens. 
Then the cost of the column must verify the following condition.
\begin{equation*}
\begin{array}{ll}
     \forall i \in [1..n],  c_i \in C_{+_{i}} \mbox {if} \\
      2 + cost(t_s) + \sum_{t_{c_j} \in C_{+_{i-1}}} cost(t_{c_j}) + cost(t_{c_i}) \leq m
\end{array}
\end{equation*}
where $C_{+_i}$ is the set of retained columns at the iteration $i$. $2$ is added to the condition as two special tokens are added to the input: $[CLS],t_s, [SEP], t{c_1},...,t{c_n}$.
If a current column $c_i$ doesn't respect the condition then the column is not selected. Whether or not the column is retained, the algorithm continues and verifies if the next column can fit. It follows $C_{+_n}$ contains the maximum number of columns that can fit under $m$ by respecting the columns scoring order.

There is a number of heuristic pruning approaches we have experimented with. Results are given in \ref{tab:apx-result-pruning-full}.

\paragraph{Word2Vec (W2V)} uses a publicly available word2vec \cite{mikolov-13} model\footnote{\url{https://tfhub.dev/google/tf2-preview/gnews-swivel-20dim/1}} to extract one embedding for each token. Let $T_S$ be the set of tokens in the statement and $T_C$ the set of tokens in a column. The cosine similarity for each pair is given by $ \forall (s,c) \in T_S \times T_C $
\begin{equation*}
 f(s, c) = \left\{ 
 \begin{array}{ll}
        1 & \mbox {if s = c} \\
        0 & \mbox {if s or c are unknown}\\
        \cos(v_s, v_c) & \mbox{else}
 \end{array}
\right.
\end{equation*}
where $v_i$ represents the embedding of the token $i$.
For a given column token $c$ we define the relevance with respect to the statement as the average similarity to every token:
\begin{equation*}
f(S, c) = \operatorname{avg}_{s \in T_S: f(s, c) > \tau}f(s, c)
\end{equation*}
Where $\tau$ is a threshold that helps to remove noise from unrelated word embeddings. We set $\tau$ to $0.89$.
We experimented with $\max$ and $\operatorname{sum}$ as other aggregation function but found the average to perform best.
The final score between the statement $S$ and the column $C$ is given by 
\begin{equation*}
f(S, C) = \max_{c\in T_C} f(S, c)
\end{equation*}

\paragraph{Term frequency–inverse document frequency (IWF)} Scores the columns' tokens proportional to the word frequency in the statement and offset by the word frequency computed over all the tables and statements from the training set.
\begin{equation*}
f(t_s, c) = \frac{TF(t_s, c)}{ \log(WF(c) + 1)}
\end{equation*}
Where $TF(t_s, c)$ is how often the token $c$ occurs in the statement $t_s$, and $WF(c)$ is the frequency of $c$ in a word count list.
The final score of a column $C$ is given by 
\begin{equation*}
f(t_s, C) = \max_{c \in T_C} \left(\frac{TF(t_s, c)}{ \log(WF(c) + 1)}\right)
\end{equation*}

\paragraph{Character N-gram (CHAR)} Scores columns by character overlap with the statement. This method looks for sub-list of word’s characters in the statement. The length of the characters' list has a minimum and maximum length allowed. In the experiments we use $5$ and $20$ as minimum and maximum length. Let $\mathbb{L}_{s,c}$ be the set of all the overlapping characters' lengths.  
The scoring for each column is given by 
\begin{equation*}
f(t_s, t_c) = \frac{min(max(\mathbb{L}_{s,c},5)),20)}{cost(t_c)}
\end{equation*}

\begin{table}[H]
\centering
\pruningtable{full}
\label{tab:apx-result-pruning-full}
\end{table}

\section{\sqa{}}
\label{sec:sqa_dev_results}

\begin{table*}[tb]
\begin{center}
\resizebox{2.0\columnwidth}{!}{
\begin{tabular}{llcccccccccc}
\toprule
 & & \multicolumn{2}{c}{\textbf{ALL}} & \multicolumn{2}{c}{\textbf{SEQ}} & \multicolumn{2}{c}{\textbf{Q1}} & \multicolumn{2}{c}{\textbf{Q2}} & \multicolumn{2}{c}{\textbf{Q3}} \\
\textbf{Data} & \textbf{Size} & Dev & Test & Dev & Test & Dev & Test & Dev & Test & Dev & Test \\
\midrule
\masklm{}                        & Base  & 60.0\err{0.3} & 64.0\err{0.2} & 35.3\err{0.7} & 34.6\err{0.0} & 72.4\err{0.4} & 79.2\err{0.6} & 59.7\err{0.4} & 61.2\err{0.4} & 50.5\err{1.1} & 55.6\err{0.7} \\
Counterfactual                   & Base  & 63.2\err{0.7} & 65.0\err{0.5} & 39.3\err{0.6} & 36.5\err{0.6} & 74.7\err{0.3} & 78.4\err{0.4} & 63.8\err{1.2} & 63.7\err{0.3} & 52.4\err{0.7} & 57.5\err{0.7} \\
Synthetic                        & Base  & 64.1\err{0.4} & 67.4\err{0.2} & 41.6\err{0.8} & 39.8\err{0.4} & 75.3\err{0.7} & 79.3\err{0.1} & 64.4\err{0.6} & 66.2\err{0.2} & 55.8\err{0.7} & 60.2\err{0.6} \\
Counterfactual + Synthetic       & Base  & 64.5\err{0.2} & 67.9\err{0.3} & 40.2\err{0.4} & 40.5\err{0.7} & 75.6\err{0.3} & 79.3\err{0.3} & 65.3\err{0.6} & 67.0\err{0.3} & 55.4\err{0.5} & 61.1\err{0.9} \\
\midrule
Counterfactual + Synthetic       & Large & 68.0\err{0.2} & 71.0\err{0.4} & 45.8\err{0.3} & 44.8\err{0.8} & 77.7\err{0.6} & 80.9\err{0.5} & 68.8\err{0.4} & 70.6\err{0.3} & 59.6\err{0.5} & 64.0\err{0.3} \\
\bottomrule
\end{tabular}}
\end{center}
\caption{\sqa{} dev (first fold) and test results. ALL is the average question accuracy, SEQ the sequence accuracy, and QX, the accuracy of the X'th question in a sequence. We show the median over 9 trials, and errors are estimated with half the interquartile range .}
\label{tab:sqa_dev_results}
\end{table*}

Table \ref{tab:sqa_dev_results} shows the accuracy on the first development fold and the test set. As for the main results, the error margins displayed are half the interquartile range over 9 runs, which is half the difference between the first and third quartile. This range contains half of the runs and provides a measure of robustness.

\section{Pre-Training Data}
\label{sec:apx-table-pruning}

When training on the pre-training data we hold out approximately 1\% of the data for testing how well the model is solving the pre-training task.
In Table \ref{tab:apx-result-data}, we compare the test pre-training accuracy on synthetic and counterfactual data to models that are only trained on the statements to understand whether there is considerable bias in the data.
Both datasets have some bias (i.e. the accuracy without table is higher than 50\%.). Still there is a sufficient enough gap between training with and without tables so that the data is still useful. 

The synthetic data can be solved almost perfectly whereas for the counterfactual data we only reach an accuracy of 84.3\%.
This is expected as there is no guarantee that the model has enough information to decide whether a statement is true or false for the counterfactual examples.

\begin{table}[H]
\centering
\resizebox{1\columnwidth}{!}{
\begin{tabular}{llccc} 
\toprule
\textbf{Data} & \textbf{Model} & \textbf{PT Size} & \multicolumn{1}{l}{\textbf{Val$_{S}$}} & \multicolumn{1}{l}{\textbf{Val$_{C}$}} \\ 
\midrule
Counterfactual           & base & 128 &       & 82.0 \\
Counterfactual w/o table  & base & 128 &       & 76.0 \\
\midrule
Synthetic             & base & 128 & 94.3 \\
Synthetic w/o table    & base & 128 & 77.8 \\
\midrule
Synthetic + Counterfactual & base & 128 &  93.7 & 79.3 \\
                        & base & 256 &  98.0 & 83.9 \\
                        & base & 512 &  98.4 & 84.3 \\
\midrule
Synthetic + Counterfactual & large & 128 & 94.3 & 81.0 \\
                        & large & 256 & 98.5 & 86.8 \\
                        & large & 512 & 98.9 & 87.3 \\
\bottomrule
\end{tabular}}
\caption{Accuracy on synthetic (Val$_{S}$) and counterfactual held-out sets (Val$_{C}$) of the pre-traininig data.}
\label{tab:apx-result-data}
\end{table}

In table~\ref{tab:apx-supported} we show the ablation results when removing the counterfactual statements that lack a supporting entity, that is a second entity that appears in both the table and sentence. This increases the probability that our generated negative pairs are incorrect, but it also discards 7 out of 8 examples, which ends up hurting the results.

\begin{table}[H]
\centering
\resizebox{1\columnwidth}{!}{
\begin{tabular}{llccc} 
\toprule
\textbf{Data} & \textbf{Val}\\ 
Synthetic                           & 77.6 \\
\midrule
Counterfactual                      & 75.5 \\
Counterfactual + Synthetic          & 78.6 \\
\midrule
Counterfactual (only supported)              & 73.6 \\
Counterfactual (only supported) + Synthetic  & 77.1 \\
\bottomrule
\end{tabular}}
\caption{Comparisons of training on counterfactual data with and without statements that don't have support mentions.}
\label{tab:apx-supported}
\end{table}

\section{Salient Groups Definition}
\label{sec:slice_words}

In table~\ref{tab:slice_words} we show the words that are used as markers to define each of the groups. We first identified manually the operations that were most often needed to solve the task and found relevant words linked with each group. The heuristic was validated by manually inspecting 50 samples from each group and observing higher than 90\% accuracy.

\begin{table}[H]
\centering
\resizebox{1.0\columnwidth}{!}{
\begin{tabular}{l|l}
\toprule													
\textbf{Slice}	&	\textbf{Words}	\\
\midrule													
\textbf{Aggregations} & total, count, average, sum,\\& amount, there, only \\
\textbf{Superlatives} & first, highest, best,\\& newest, most, greatest, latest, \\& biggest and their opposites \\
\textbf{Comparatives} & than, less, more, better, \\ & worse, higher, lower, shorter, same \\
\textbf{Negations}    & not, any, none, no, never \\
\bottomrule
\end{tabular}
}
\caption{Trigger words for different groups.}
\label{tab:slice_words}
\end{table}

\end{document}